\documentclass{article}

\usepackage{arxiv}
\pdfoutput=1

\usepackage[utf8]{inputenc} 
\usepackage[T1]{fontenc}    
\usepackage{url}            
\usepackage{booktabs}       
\usepackage{amsfonts}       
\usepackage{nicefrac}       
\usepackage{microtype}      
\usepackage{lipsum}
\usepackage{enumitem}
\usepackage{natbib}
\usepackage{url}            
\usepackage{graphicx}
\usepackage{subfigure}
\usepackage[skins]{tcolorbox}
\usepackage{hyperref}
\usepackage[T1]{fontenc}
\usepackage{epigraph}
\usepackage[export]{adjustbox}
\usepackage{amssymb,mathtools}
\usepackage{multirow}
\usepackage{enumitem}
\usepackage{amsmath}
\usepackage{hyperref}
\hypersetup{
    colorlinks=False,
    linkcolor=blue,
    filecolor=magenta,      
}

\title{What Do Compressed Deep Neural Networks Forget?}

\author{
  Sara Hooker \thanks{Correspondence should be directed to shooker@google.com} \\
  Google Brain \\
  \And
  Aaron Courville \\
  MILA \\
  \And
  Gregory Clark \\
  Google \\
  \And
  Yann Dauphin \\
  Google Brain \\
  \And
  Andrea Frome \\
  Google Brain \\
}

\begin{document}

\maketitle

\begin{abstract}

Deep neural network pruning and quantization techniques have demonstrated it is possible to achieve high levels of compression with surprisingly little degradation to test set accuracy. However, this measure of performance conceals significant differences in how different classes and images are impacted by model compression techniques. We find that models with radically different numbers of weights have comparable top-line performance metrics but diverge considerably in behavior on a narrow subset of the dataset. This small subset of data points, which we term Pruning Identified Exemplars (PIEs), are systematically more impacted by the introduction of sparsity. Our work is the first to provide a formal framework for auditing the disparate harm incurred by compression and a way to quantify the trade-offs involved. An understanding of this disparate impact is critical given the widespread deployment of compressed models in the wild.

\end{abstract}

\section{Introduction} \label{submission}

Between infancy and adulthood, the number of synapses in our brain first multiply and then fall. Synaptic pruning improves efficiency by removing redundant neurons and strengthening synaptic connections that are most useful for the environment \citep{RAKI1994}. Despite losing $50 \%$ of all synapses between age two and ten, the brain continues to function \citep{kolb2009fundamentals,Sowell8223}. The phrase "Use it or lose it" is frequently used to describe the environmental influence of the learning process on synaptic  pruning, however there is little scientific consensus on \emph{what} exactly is lost \citep{CASEY2000241}.

In this work, we ask what is \textit{lost} when we compress a deep neural network. Work since the 1990s has shown that deep neural networks can be pruned of ``excess capacity'' in a similar fashion to synaptic pruning \citep{Cun90optimalbrain,Hassibi93secondorder,1992_nowlan_hinton, NIPS1990_Andreas_weight_elimination}. At face value, compression appears to promise you can have it all. Deep neural networks are remarkably tolerant of high levels of pruning and quantization with an almost negligible loss to top-1 accuracy \citep{lwac, sws, 2017Liu, 2017l0_reg, 2014memorybounded, namhoon2018}. These more compact networks are frequently favored in resource constrained settings; compressed models require less memory, energy consumption and have lower inference latency \citep{Reagen_7551399, 7551407Chen, fisher-pruning, wavernn, lpcnet,tessera2021}. 

The ability to compress networks with seemingly so little degradation to generalization performance is puzzling. How can networks with radically different representations and number of parameters have comparable top-level metrics? One possibility is that test-set accuracy is simply not a precise enough measure to capture how compression impacts the generalization properties of the model. Despite the widespread use of compression techniques, articulating the trade-offs of compression has overwhelmingly focused on change to overall top-1 accuracy for a given level of compression. 

The cost to top-1 accuracy appears minimal if it is spread uniformally across all classes, but what if the cost is concentrated in only a few classes? \textit{Are certain types of examples or classes disproportionately impacted by compression?} In this work, we propose a formal framework to audit the impact of compression on generalization properties beyond top-line metrics. Our work is the first to our knowledge that asks how dis-aggregated measures of model performance at a class and exemplar level are impacted by compression.

\textbf{Contributions}  We run thousands of large scale experiments and establish consistent results across multiple datasets--- CIFAR-10 \citep{Krizhevsky09learningmultiple}, CelebA \citep{liu2015faceattributes} and ImageNet \citep{imagenet_cvpr09}, widely used pruning and quantization techniques, and model architectures. We find that: 
\begin{enumerate}
\itemsep0em
\item  Top-line metrics such as top-1 or top-5 test-set accuracy hide critical details in the ways that pruning impacts model generalization. Certain parts of the data distribution are far more sensitive to varying the number of weights in a network, and bear the brunt of the cost of varying the weight representation.
\item  The examples most impacted by pruning, which we term \textit{Pruning Identified Exemplars (PIEs)}, are more challenging for both models and humans to classify. We conduct a human study and find that PIEs tend to be mislabelled, of lower quality, depict multiple objects, or require fine-grained classification. Compression impairs the model's ability to predict accurately on the long-tail of less frequent instances.
\item  Pruned networks are more sensitive to natural adversarial images and corruptions. This sensitivity is amplified at higher levels of compression.
\item While all compression techniques that we evaluate have a non-uniform impact, not all methods are created equal. High levels of pruning incur a far higher disparate impact than is observed for the quantization techniques that we evaluate. 
\end{enumerate}



\begin{figure}
\setlength\tabcolsep{3pt}
\begin{tabular}{cccccccc}
\toprule
 \multicolumn{2}{c}{\fontfamily{bch}\selectfont \textbf{toilet seat}} & 
 \multicolumn{2}{c}{\fontfamily{bch} \selectfont \textbf{espresso}} &

\multicolumn{2}{c}{\fontfamily{bch}\selectfont \textbf{plastic bag}} \\
 \includegraphics[width=1.in]{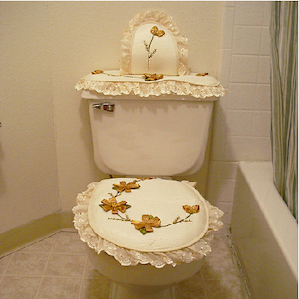} & 
\includegraphics[width=1.in]{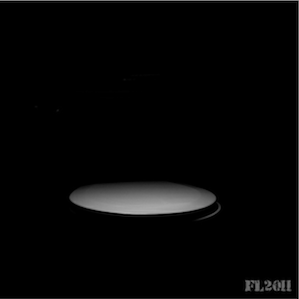} &
\includegraphics[width=1.in]{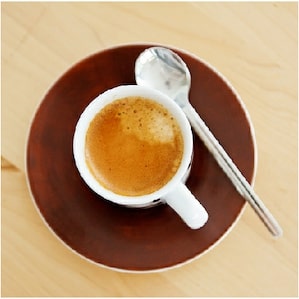} & 
\includegraphics[width=1.in]{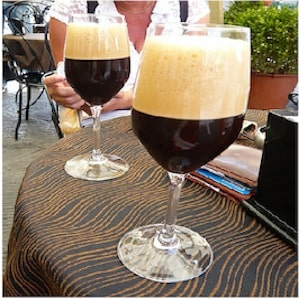} &
\includegraphics[width=1.in]{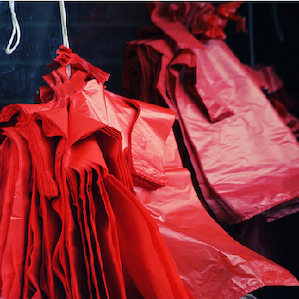} &
\includegraphics[width=1.in]{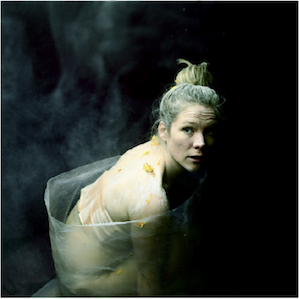} \\
 \addlinespace
{\fontfamily{qag}\selectfont \small Non-PIE} &
 {\fontfamily{qag}\selectfont \small PIE}  &
 {\fontfamily{qag}\selectfont \small Non-PIE} &
 {\fontfamily{qag}\selectfont \small PIE}   &
 {\fontfamily{qag}\selectfont \small Non-PIE}  & 
 {\fontfamily{qag}\selectfont \small PIE} \\
  \addlinespace
 \midrule
\multicolumn{2}{c}{\fontfamily{bch}\selectfont \textbf{matchstick}} &
 \multicolumn{2}{c}{\fontfamily{bch} \selectfont \textbf{cloak}}
 &
\multicolumn{2}{c}{\fontfamily{bch}\selectfont \textbf{stretcher}} \\
\includegraphics[valign=m,width=1.in]{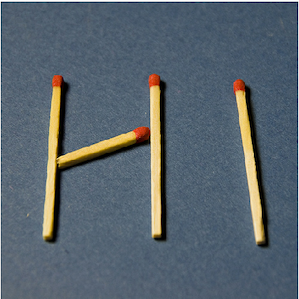} &
 \includegraphics[valign=m,width=1.in]{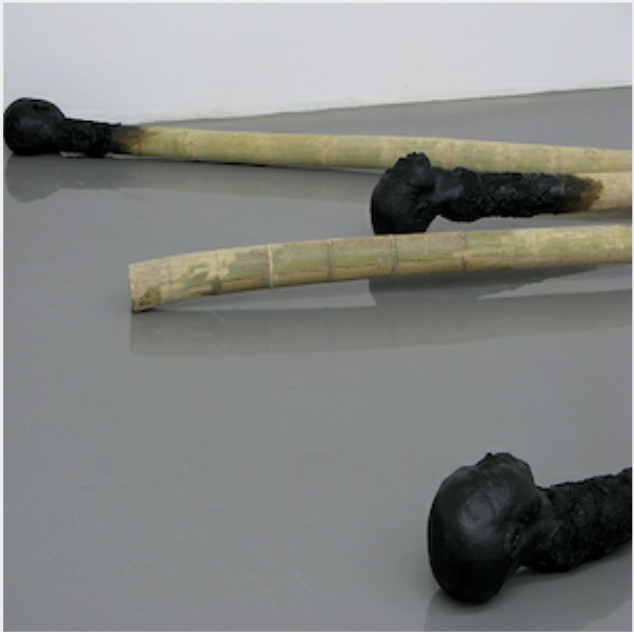} &
 \includegraphics[valign=m,width=1.in]{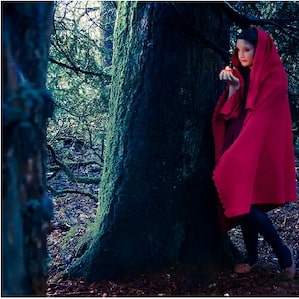} &
 \includegraphics[valign=m,width=1.in]{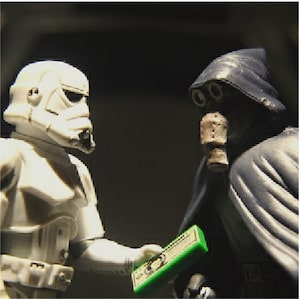} &
\includegraphics[valign=m,width=1.in]{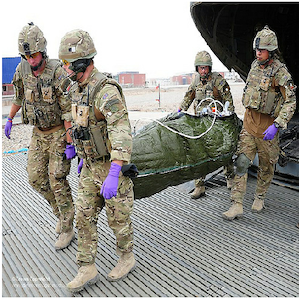} &
 \includegraphics[valign=m,width=1.in]{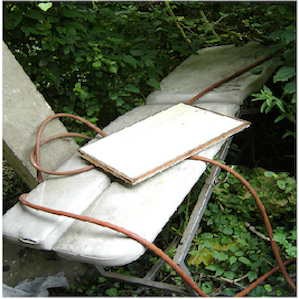} 
\\
 \addlinespace
 {\fontfamily{qag}\selectfont \small Non-PIE} &
 {\fontfamily{qag}\selectfont \small PIE}  &
 {\fontfamily{qag}\selectfont \small Non-PIE} &
 {\fontfamily{qag}\selectfont \small PIE} &
 {\fontfamily{qag}\selectfont \small Non-PIE} &
 {\fontfamily{qag}\selectfont \small PIE} \\
 \addlinespace
 \midrule
 \multicolumn{2}{c}{\fontfamily{bch}\selectfont \textbf{wool}} 
 &
 \multicolumn{2}{c}{\fontfamily{bch} \selectfont \textbf{maze}} &
\multicolumn{2}{c}{\fontfamily{bch}\selectfont \textbf{gas pump}} \\
 \includegraphics[width=1.in]{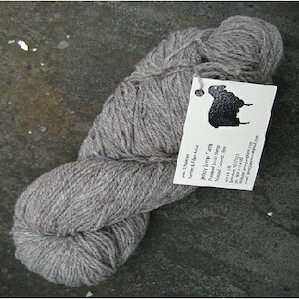} & 
\includegraphics[width=1.in]{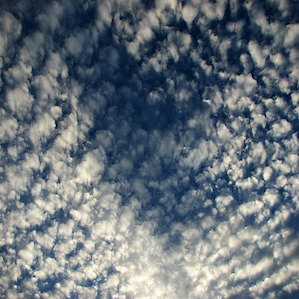} &
\includegraphics[width=1.in]{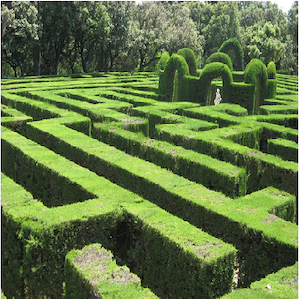} & 
\includegraphics[width=1.in]{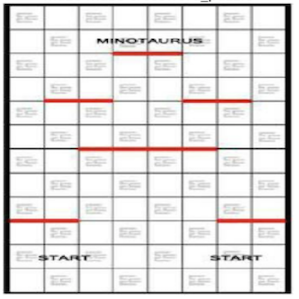} &
\includegraphics[width=1.in]{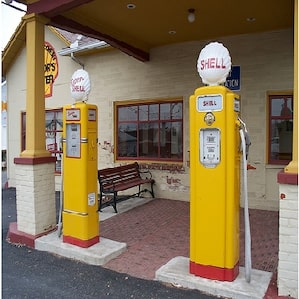} &
\includegraphics[width=1.in]{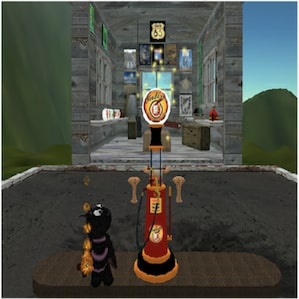} \\
 \addlinespace
{\fontfamily{qag}\selectfont \small Non-PIE} &
 {\fontfamily{qag}\selectfont \small PIE} & 
 {\fontfamily{qag}\selectfont \small Non-PIE} &
 {\fontfamily{qag}\selectfont \small PIE} & 
 {\fontfamily{qag}\selectfont \small Non-PIE}  & 
 {\fontfamily{qag}\selectfont \small PIE} \\
\end{tabular}
\caption{Pruning Identified Exemplars (PIEs) are images where there is a high level of disagreement between the predictions of pruned and non-pruned models. Visualized are a sample of ImageNet PIEs alongside a non-PIE image from the same class. Above each image pair is the true label.}
\label{fig:pie_exemplars}
\end{figure}

Our work provides intuition into the role of capacity in deep neural networks and a mechanism to audit the trade-offs incurred by compression. Our findings suggest that caution should be used before deploying compressed networks to sensitive domains. Our PIE methodology could conceivably be explored as a mechanism to surface a tractable subset of atypical examples for further human inspection \citep{Leibig2017,ZHANG1992}, to choose not to classify certain examples when the model is uncertain \citep{Bartlett2008, NIPS2016Cortes}, or to aid interpretability as a case based reasoning tool to explain model behavior \citep{NIPS2016_6300, Caruana2000, Hooker2019ABF}.

\section{Methodology and Experiment Framework}\label{section:methodology}

\subsection{Preliminaries}
We consider a supervised classification problem where a deep neural network is trained to approximate the function $F$ that maps an input variable $X$ to an output variable $Y$, formally $F:X \mapsto Y$. The model is trained on a training set of $N$ images $\mathcal{D} = \left\{(x_i, y_i)\right\}^N_{i=1}$, and at test time makes a prediction $y_i^*$ for each image in the test set. The true labels $y_i$ are each assumed to be one of $C$ classes, such that $y_i = [1,....,C]$. 

A reasonable response to our desire for more compact representations is to simply train a network with fewer weights. However, as of yet, starting out with a compact dense model has not yielded competitive test-set performance \citep{li2020train,zhu2017prune}. Instead, research has centered on a more tractable direction of investigation -- the model begins training with "excess capacity" and the goal is to remove the parts that are not strictly necessary for the task by/at the end of training. A pruning method $\mathcal{P}$ identifies the subset of weights to set to zero. A sparse model function, $\hat{f}_{t}^p$, is one where a fraction $t$ of all model weights are set to zero. Equating weight value to zero effectively removes the contribution of a weight, as multiplication with inputs no longer contributes to the activation.  A non-compressed model function is one where all weights are trainable ($t=0$). We refer to the overall model accuracy as $\beta_t^\mathcal{M}$. In contrast, $t=0.9$ indicates that $90\%$ of model weights are removed over the course of training, leaving a maximum of $10\%$ non-zero weights.

\subsection{Class level measure of impact}

If the impact of compression was completely uniform, the relative relationship between class level accuracy  $\beta_t^c$ and overall model performance will be unaltered. This forms our null hypothesis (\textbf{$H_0$}). We must decide for each class $c$ whether to reject the null hypothesis and accept the alternate hypothesis (\textbf{$H_1$}) - the relative change to class level recall differs from the change to overall accuracy in either a positive or negative direction:
\begin{align}
H_0: \frac{\beta_0^c}{\beta_0^\mathcal{M}} = \frac{\beta_t^c}{\beta_t^\mathcal{M}}\\ 
H_1: \frac{\beta_0^c}{ \beta_0^\mathcal{M}} \neq \frac{\beta_t^c}{\beta_t^\mathcal{M}}
\end{align}

\textbf{Welch's t-test} Evaluating whether the difference between the samples of mean-shifted class accuracy from compressed and non-compressed models is ``real'' amounts to determining whether these two data samples are drawn from the same underlying distribution, which is the subject of a large body of goodness of fit literature \citep{1986agostino, 1954anderson_darling, 2002huber}. We independently train a population of $K$ models for each compression method, dataset, and model that we consider. Thus, we have a sample $S_t^c$ of accuracy metrics per class $c$ at each level of compression $t$. 

For each class $c$, we we use a two-tailed, independent Welch's t-test \citep{1947welch} to determine whether the mean-shifted class accuracy $S_t^c = \{\beta_{t,k}^c - \beta_{t,k}^\mathcal{M} \}_{k=1}^K$ of the samples $S_t^c$ and $S_0^c$ differ significantly. If the $p$-value $<= 0.05$, we reject the null hypothesis and consider the class to be disparately impacted by $t$ level of compression relative to the baseline.

\textbf{Controlling for overall changes to top-line metrics} Note that by comparing the relative difference in  class accuracy $S_t^c$, we control for any overall difference in model test-set accuracy. This is important because while small, the difference in top-line metrics is not zero (see Table.~\ref{table:imagenet_summary}). Along with the $p$-value, for each class we report the average relative deviation in class-level accuracy, which we refer to as {\em relative recall difference}:
\begin{equation}
\frac{1}{K} \sum_{k=1}^{K} \left(
\frac{\beta_{t,k}^c}{\beta_{0,k}^c}\right) 
\label{eq:norm_diff}
\end{equation}

\begin{table*}[ht!]
\centering
    \begin{tabular}{ccc}
    \includegraphics[width=0.3\textwidth]{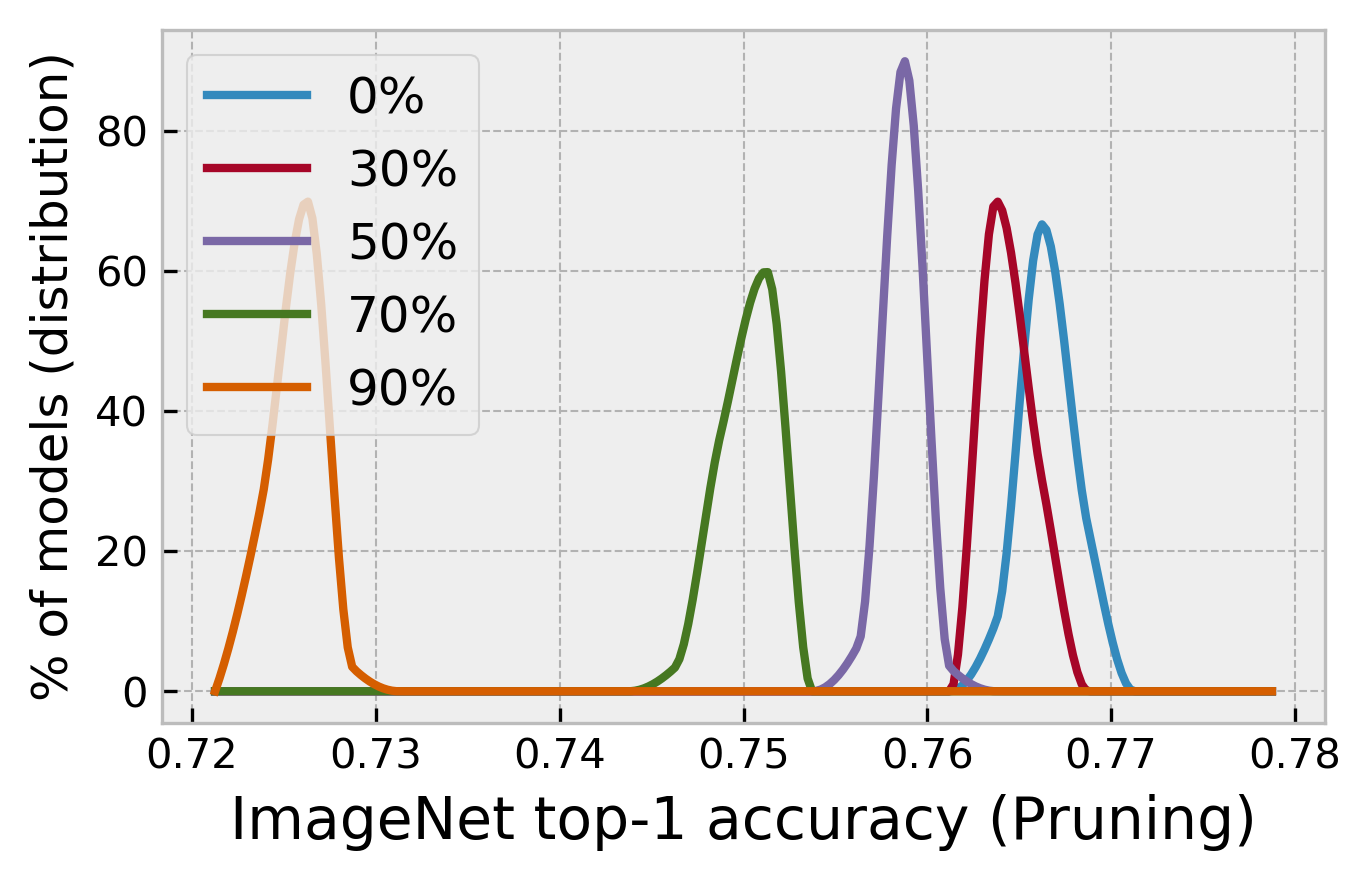} &
    \includegraphics[width=0.3\textwidth]{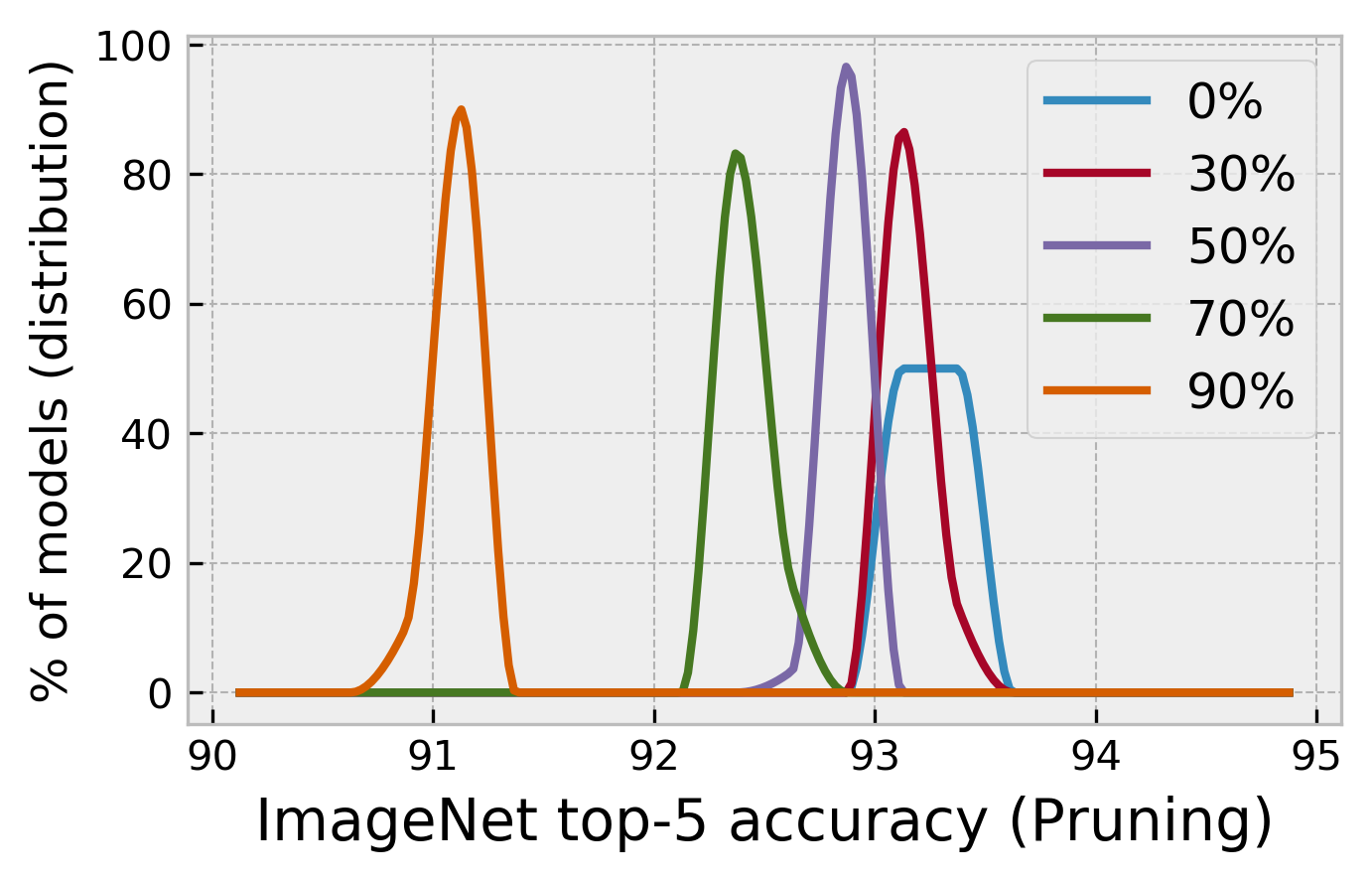} &
    \includegraphics[width=0.3\textwidth]{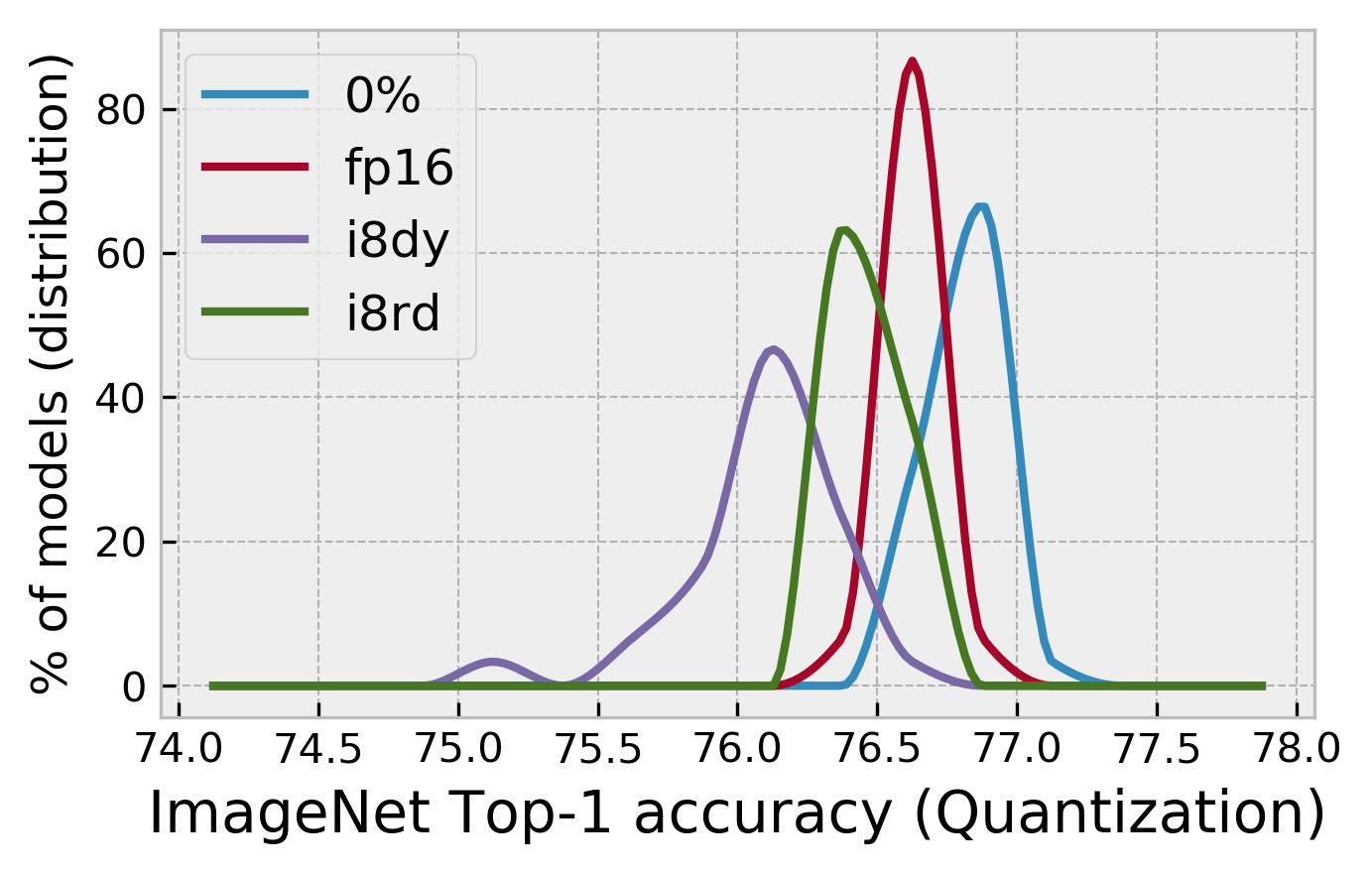} \\ 
    \includegraphics[width=0.3\textwidth]{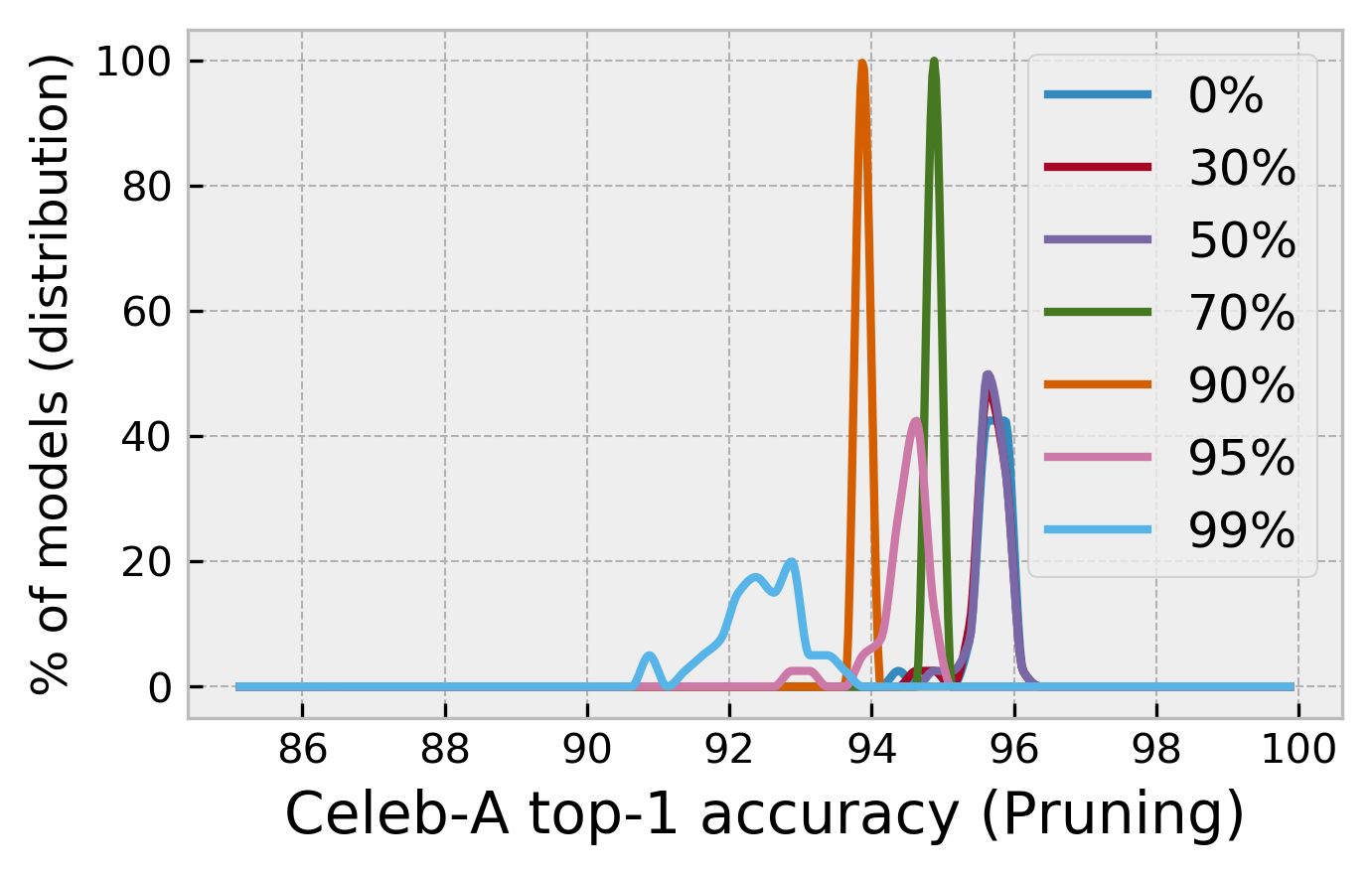} &
    \includegraphics[width=0.3\textwidth]{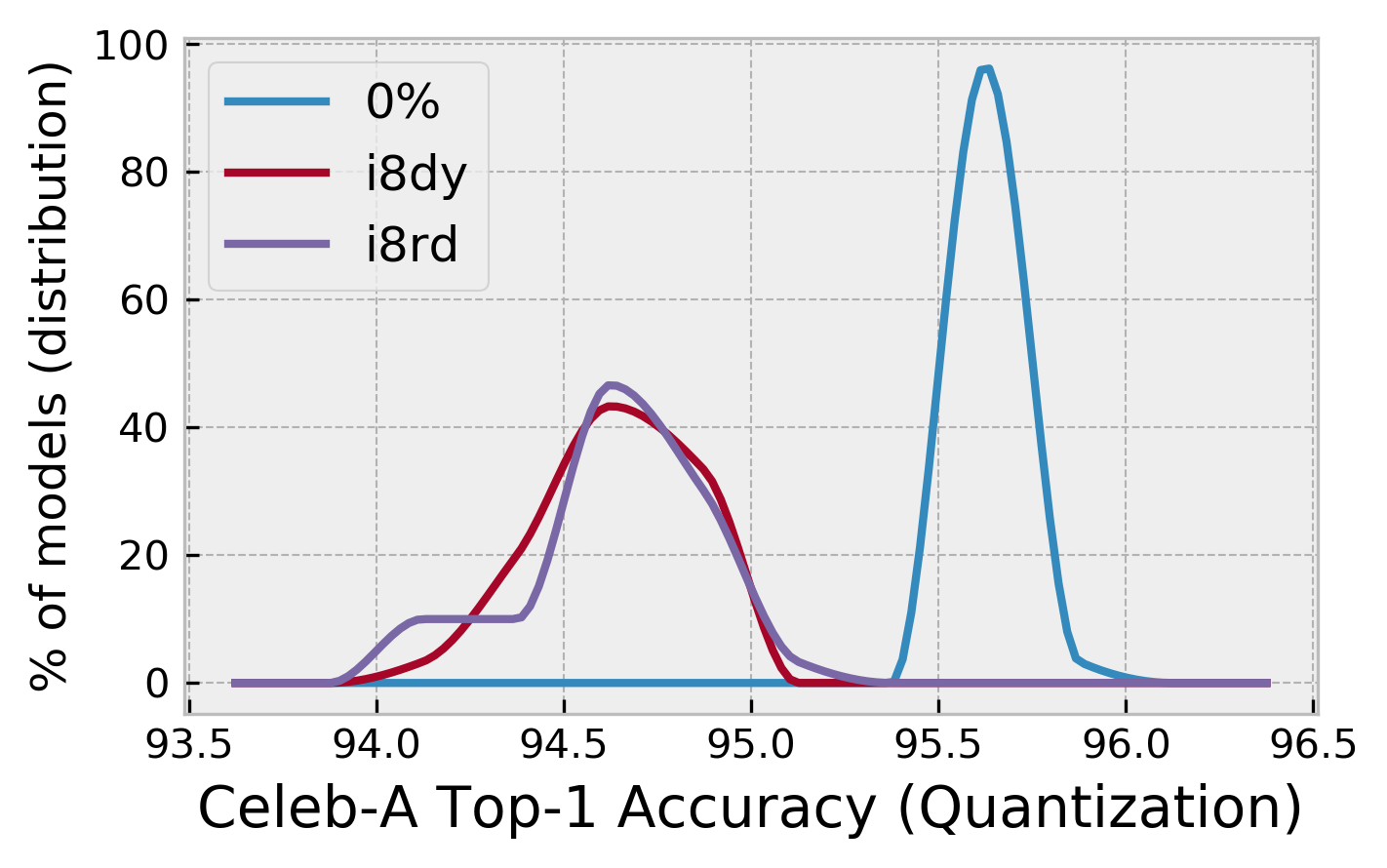} &
    \includegraphics[width=0.3\textwidth]{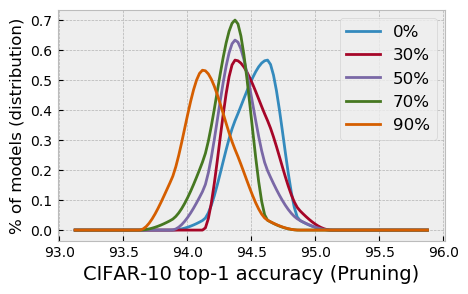}
    \end{tabular}
    \caption{Distributions of top-1 accuracy for populations of independently quantized and pruned models for ImageNet, CIFAR-10 and CelebA. For ImageNet, we also include top-5. Note that the scale of the x-axis differs between plots.}
    \label{fig:modeldist_all}
\end{table*}

\subsection{Pruning Identified Exemplars}

In addition to measuring the class level impact of compression, we are interested in how model predictive behavior changes through the compression process. Given the limitations of un-calibrated probabilities in deep neural networks \citep{2017Guo, NIPSKendall2017}, we focus on the level of disagreement between the predictions of compressed and non-compressed networks on a given image. 
Using the populations of models $K$ described in the prior section, we construct sets of predictions $Y^*_{i,t} = \{ y^*_{i,k,t} \}_{k=1}^K$ for a given image $i$. 

For set $Y^*_{i,t}$ we find the \textit{modal label}, i.e. the class predicted most frequently by the $t$-pruned model population for image $i$, which we denote $y^M_{i,t}$. The exemplar is classified as a pruning identified exemplar $\textit{PIE}_t$ if and only if the modal label is different between the set of $t$-pruned models and the non-pruned baseline models:
\[
\textit{PIE}_{i,t} = 
\begin{cases}
1 & \text{if $y^M_{i,0} \neq y^M_{i,t}$} \\
0 & \text{otherwise}
\end{cases}
\]
We note that there is no constraint that the non-pruned predictions for PIEs match the true label. Thus the detection of PIEs is an unsupervised protocol that can be performed at test time. 

\subsection{Experimental framework}

\textbf{Tasks}  We evaluate the impact of compression across three classification tasks and models: a wide ResNet model \citep{Zagoruyko2016} trained on CIFAR-10, a ResNet-50 \citep{He_2015} trained on ImageNet, and a ResNet-18 trained on CelebA. All networks are trained with batch normalization \citep{Ioffe2015}, weight decay, decreasing learning rate schedules, and augmented training data. We train for $32,000$ steps (approximately $90$ epochs) on ImageNet with a batch size of $1024$ images, for $80,000$ steps on CIFAR-10 with a batch size of $128$, and $10,000$ steps on CelebA with a batch size of $256$. For ImageNet, CIFAR-10 and CelebA, the baseline non-compressed model obtains a mean top-1 accuracy of $76.68\%$,  $94.35\%$ and  $94.73\%$ respectively. Our goal is to move beyond anecdotal observations, and to measure statistical deviations between populations of models. Thus, we report metrics and statistical significance for each dataset, model and compression variant across $30$ independent trainings.

\begin{figure*}[ht!]
    \centering
    \begin{sc}
    \begin{tabular}[t]{cc}
    \includegraphics[width=.45\columnwidth,height=10cm,keepaspectratio]{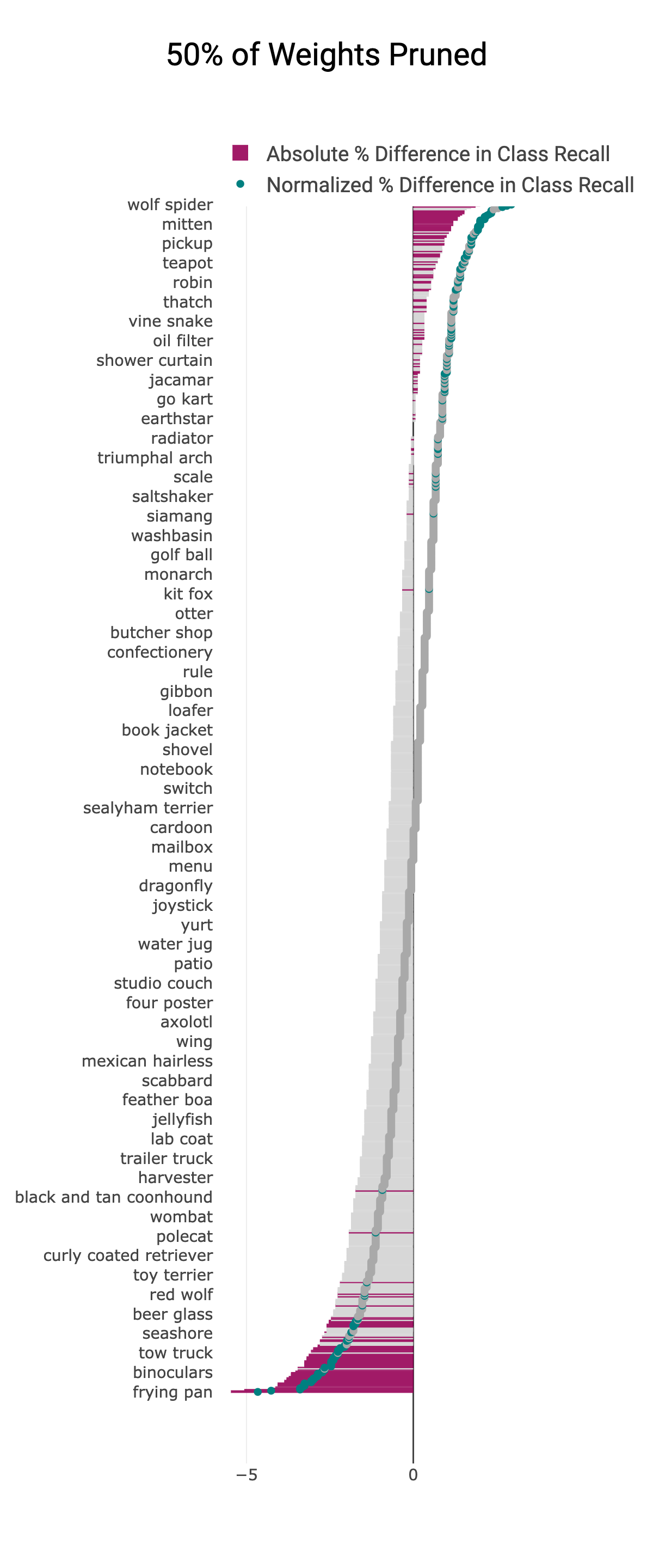} &
    \includegraphics[width=.45\columnwidth,height=10cm,keepaspectratio]{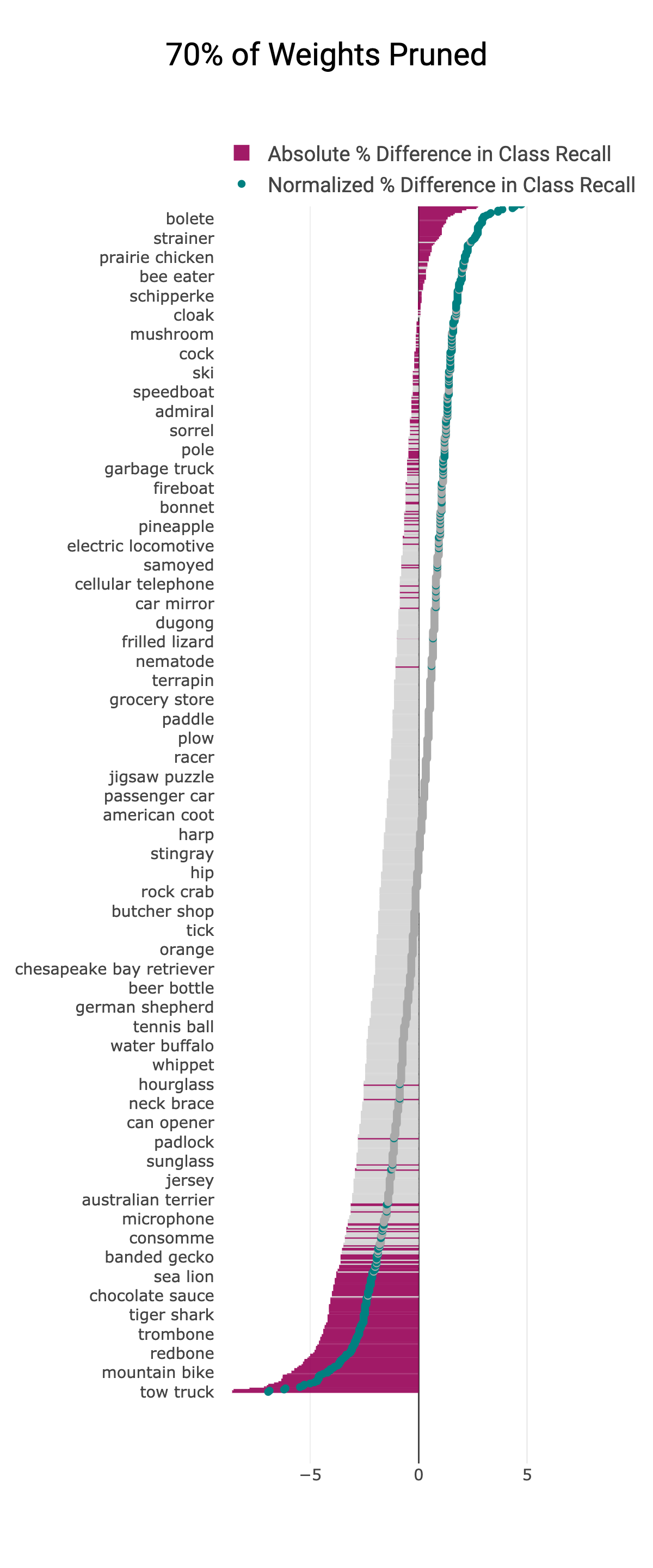}
    \includegraphics[width=.45\columnwidth,height=10cm,keepaspectratio]{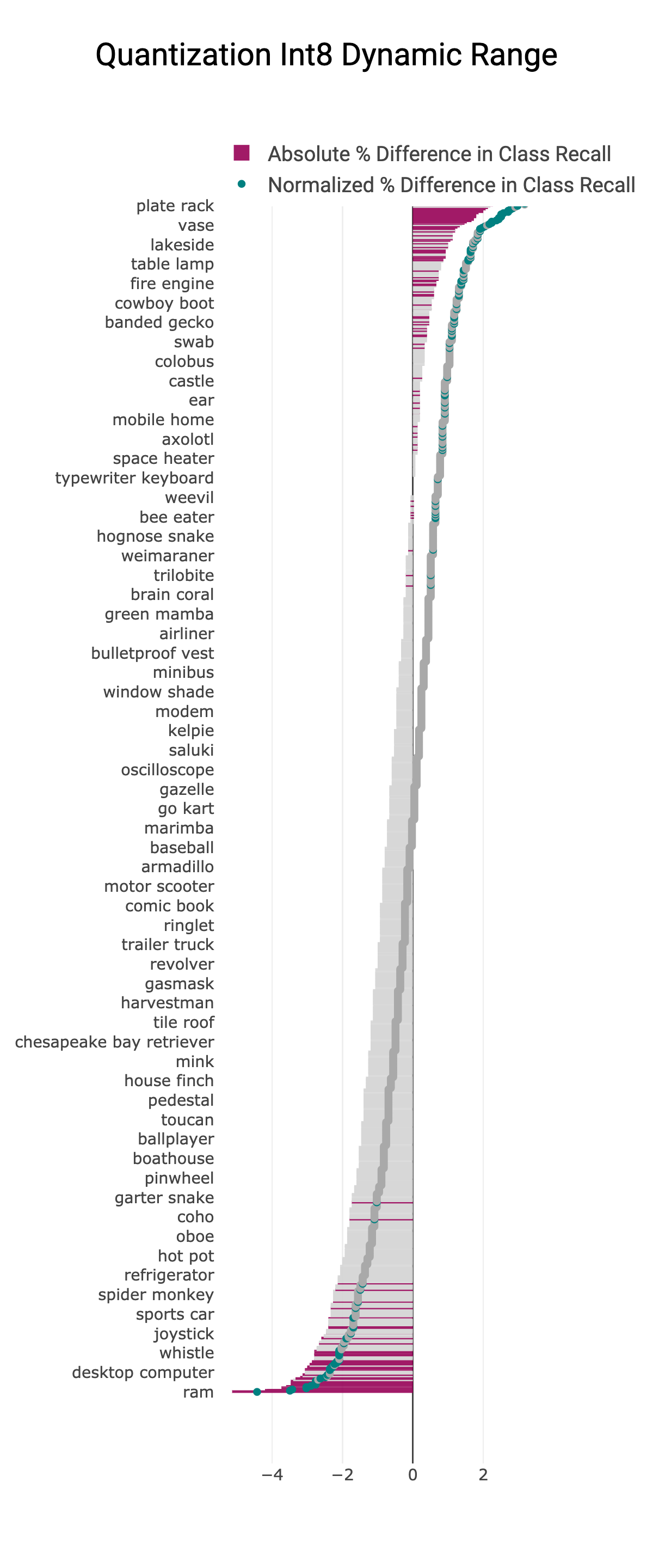}
    \end{tabular}
    \end{sc}
    \caption{Compression disproportionately impacts a small subset of ImageNet classes. Plum bars indicate the subset of examples where the impact of compression is statistically significant. Green scatter points show normalized recall difference which normalizes by overall change in model accuracy, and the bars show absolute recall difference. \textbf{Left:} $50\%$ pruning. Center: $70\%$ pruning. \textbf{Right:} post-training int8 dynamic range quantization. The class labels are sampled for readability.}
    \label{fig:imagenet_recall_70_90_quantized}
\end{figure*}

\textbf{Pruning and quantization techniques considered}  We evaluate magnitude pruning as proposed by \citet{Gupta2017}. For pruning, we vary the end sparsity precisely for $t \in \{0.3, 0.5, 0.7, 0.9\}$. For example, $t=0.9$ indicates that $90\%$ of model weights are removed over the course of training, leaving a maximum of $10\%$ non-zero weights. For each level of pruning $t$, we train $30$ models from random initialization.

We evaluate three different quantization techniques: float16 quantization \texttt{float16} \citep{2017Micikevicius}, hybrid dynamic range quantization with int8 weights \texttt{hybrid} \citep{Alvarez_2016} and fixed-point only quantization with int8 weights created with a small representative dataset \texttt{fixed-point} \citep{Vanhoucke_2011, Jacob_2018}. 

All quantization methods we evaluate are implemented post-training, in contrast to the pruning which is applied progressively over the course of training. We use a limited grid search to tailor the pruning schedule and hyperparameters to each dataset to maximize top-1 accuracy. We include additional details about training methodology and pruning techniques in the supplementary material. All the code for this paper is publicly available \href{https://github.com/google-research/google-research/tree/master/pruning_identified_exemplars}{here}.

\section{Results}\label{section:results}

\subsection{Disparate impact of compression}

We find consistent results across all datasets and compression techniques considered; a small subset of classes are disproportionately impacted. This disparate impact is far from random, with statistically significant differences in class level recall between a population of non-compressed and compressed models. Compression induces \textit{``selective forgetting''} with performance on certain classes evidencing far more sensitivity to varying the representation of the network. This sensitivity is amplified at higher levels of sparsity with more classes evidencing a statistically significant relative change in recall. For example, as seen in Table \ref{table:imagenet_summary} at $50\%$ sparsity 170 ImageNet classes are statistically significant which increases to $372$ classes at $70\%$ sparsity. 

\textbf{Cannibalizing a small subset of classes} Out of the classes where there is a statistically significant deviation in performance, we observe a subset of classes that benefit relative to the average class as well as classes that are impacted adversely. However, the average absolute class decrease in recall is far larger than the average increase, meaning that the losses in generalization caused by pruning is far more concentrated than the relative gains. Compression cannibalizes performance on a small subset of classes to preserve a similar overall top-line accuracy.

\textbf{Comparison of quantization and pruning techniques} While all the techniques we benchmark evidence disparate class level impact, we note that quantization appears to introduce less disparate harm. For example, the most aggressive form of post-training quantization considered, fixed-point only quantization with int8 weights \texttt{fixed-point}, impacts the {\em relative recall difference} of 119 ImageNet classes in a statistically significant way. In contrast, at $90 \%$ sparsity, {\em relative recall difference} is statistically significant for 637 classes. These results suggest that the representation learnt by a network is far more robust to changes in precision versus removing the weights entirely. For sensitive tasks,  quantization may be more viable for practitioners as there is less systematic disparate impact.

\textbf{Complexity of task} The impact of compression depends upon the degree of overparameterization present in the network given the complexity of the task in question. For example, the ratio of classes that are significantly impacted by pruning was lower for CIFAR-10 than for ImageNet. One class out of ten was significantly impacted at $30\%$ and $50\%$, and two classes were impacted at $90\%$. We suspect that we measured less disparate impact for CIFAR-10 because, while the model has less capacity, the number of weights is still sufficient to model the limited number of classes and lower dimensional dataset. In the next section, we leverage PIEs to characterize and gain intuition into why certain parts of the distribution are systematically far more sensitive to compression.

\subsection{Pruning Identified Exemplars}

To better understand why a narrow part of the data distributon is far more sensitive to compression, we ($1$) evaluate whether PIEs are more difficult for an algorithm to classify, ($2$) conduct a human study to codify the attributes of a sample of PIEs and Non-PIEs, and ($3$) evaluate whether PIEs over-index on underrepresented sensitive attributes in CelebA.

At every level of compression, we identify a subset of PIE images that are disproportionately sensitive to the removal of weights (for each of CIFAR-10, CelebA and ImageNet). The number of images classified as PIE increases with the level of pruning. At $90 \%$ sparsity, we classify $10.27 \%$ of all ImageNet test-set images as PIEs, $2.16 \%$ of CIFAR-10, and $16.17 \%$ of CelebA.

\begin{figure}
    \centering
\begin{tabular}{ccc}
\toprule
\multicolumn{3}{c}{\textbf{Top-1 Accuracy on PIE, All Test-Set, Non-PIE}}  \\
\midrule
\textbf{CelebA} & \textbf{CIFAR-10} & \textbf{ImageNet} \\
\midrule
\includegraphics[width=.3\textwidth,scale=1]{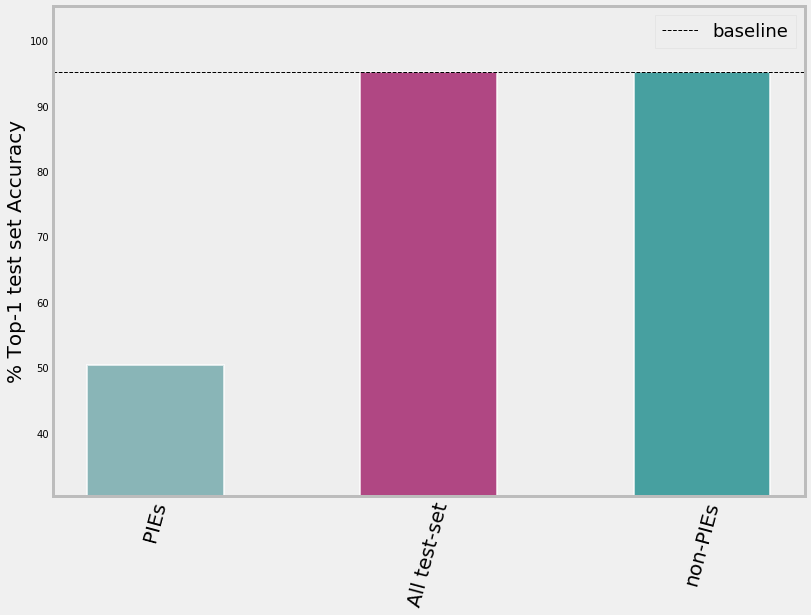} &
\includegraphics[width=.3\textwidth,scale=1]{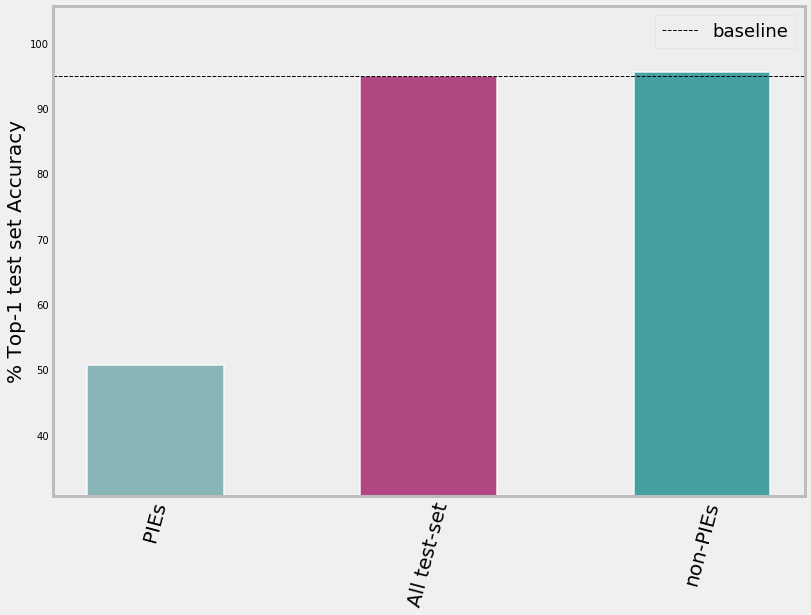} &
\includegraphics[width=.3\textwidth,scale=1]{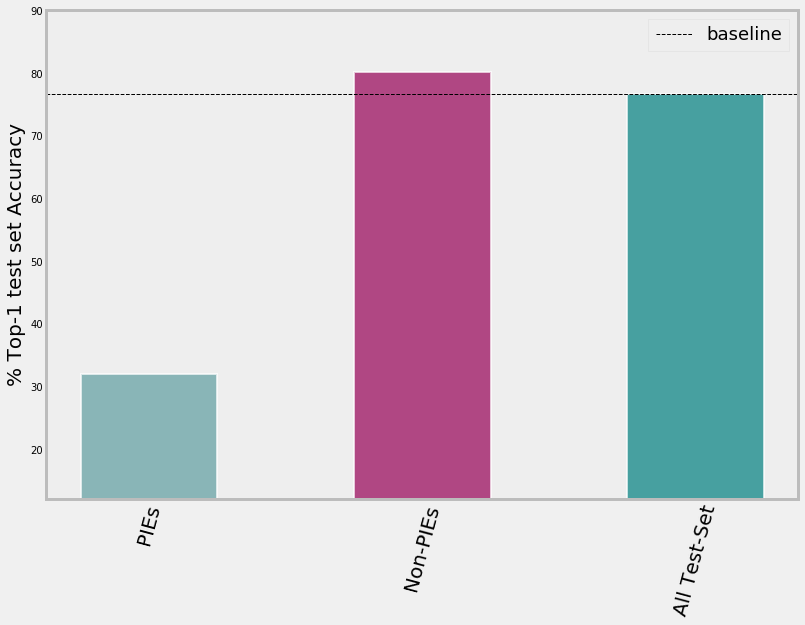} \\
\midrule
\end{tabular}
\caption{A comparison of model performance on $1$) a sample of Pruning Identified Exemplars (PIE), $2$) the entire test-set and $3$) a sample excluding PIEs. Inference on the non-PIE sample improves test-set top-1 accuracy relative to the baseline for ImageNet. Evaluation on PIE images alone yields substantially lower top-1 accuracy.}
\label{fig:inference_pie_only}
\end{figure}

\textbf{Test-error on PIEs} In Fig.~\ref{fig:inference_pie_only}, we evaluate a random sample of ($1$) PIE images, ($2$) non-PIE images and ($3$) entire test-set for each of the datasets considered. We find that PIE images are far more challenging for a non-compressed model to classify. Evaluation on PIE images alone yields substantially lower top-1 accuracy. The results are consistent across CIFAR-10 (top-1 accuracy falls from $94.89 \%$  to $43.64  \%$), CelebA ($94.10 \%$  to $50.41 \%$), and ImageNet datasets ($76.75  \%$  to $39.81  \%$). Notably, on ImageNet, we find that removing PIEs greatly improves generalization performance. Test-set accuracy on non-PIEs increased to $81.20  \%$ relative to baseline top-1 performance of $76.75 \%$.

\textbf{Human study}  We conducted a human study ($85$ participants) to label a random sample of $1230$ PIE and non-PIE ImageNet images. Humans in the study were shown a balanced sample of PIE and non-PIE images that were selected at random and shuffled. The classification as PIE or non-PIE was not known or available to the human.\textit{What makes PIEs different from non-PIEs?} The participants were asked to codify a set of attributes for each image. We report the relative distribution of PIE and non-PIE after each attribute, with the higher relative share in bold:
\begin{enumerate}
\itemsep-0.2em
  \item \textbf{ground truth label incorrect or inadequate -- } image contains insufficient information for a human to arrive at the correct ground truth label. [$8.90\%$ of non-PIEs, \textbf{20.05\%} of PIEs]
     \item  \textbf{multiple-object image} -- image depicts multiple objects where a human may consider several labels to be appropriate (e.g., an image which depicts both a \texttt{paddle} and \texttt{canoe} or a \texttt{desktop computer} consisting of a \texttt{screen}, \texttt{mouse}, and \texttt{monitor}). [$39.53\%$ of non-PIE, \textbf{59.15 \%} of PIEs]
         \item \textbf{corrupted image --} image exhibits common corruptions such as motion blur, contrast, pixelation. We also include in this category images with super-imposed text or an artificial frame as well as images that are black and white rather than the typical RGB color images in ImageNet. [\textbf{14.37\%} of non-PIE, $13.72\%$ of PIEs]
         \item  \textbf{fine grained classification --} image involves classifying an object that is semantically close to various other class categories present in the dataset (e.g., \texttt{rock crab} and \texttt{fiddler crab}, \texttt{bassinet} and \texttt{cradle}, \texttt{cuirass} and \texttt{breastplate}). [$8.9\%$ of non-PIEs, \textbf{43.55\%} of PIEs]
 \item \textbf{abstract representations --} image depicts a class object in an abstract form such a cartoon, painting, or sculptured incarnation of the object. [$3.43\%$ of non-PIE, \textbf{5.76\%} of PIE]
 \itemsep-0.2em
\end{enumerate}

\begin{table*}[t]
\begin{center}
\begin{sc}
\begin{tabular}{ccccc}
\toprule
\textbf{Fraction Pruned} & \textbf{Top 1} & \textbf{Top 5}
& \textbf{Count Signif Classes} 
& \textbf{Count PIEs} \\
\midrule
  0    & 76.68 & 93.25 & - & - \\
  30 & 76.46 & 93.17 & 68 & 1,819 \\
  50 & 75.87 & 92.86 & 170 & 2,193 \\
  70 & 75.02 & 92.43 & 372 & 3,073 \\
  90 & 72.60 & 91.10 & 637 & 5,136 \\
 \midrule
 \textbf{Quantization} &  &  &  \\
  \midrule
float16 & 76.65  & 93.25 & 58 & 2019 \\
dynamic range int8 & 76.10 & 92.94 & 144 & 2193  \\
fixed-point int8 & 76.46 & 93.16 & 119 & 2093  \\
 \bottomrule
\end{tabular}
\end{sc}
 \end{center}
 \caption{ImageNet top-1 and top-5 accuracy at all levels of pruning and quantization, averaged over all runs. Count PIEs is the count of images classified as a Pruning Identified Exemplars at every compression level. We include comparable tables for CelebA and CIFAR-10 in the appendix.}
\label{table:imagenet_summary} 
\end{table*}

PIEs heavily over-index relative to non-PIEs on certain properties, such as having an \textit{incorrect ground truth label}, involving a \textit{fine-grained classification task} or \textit{multiple objects}. This suggests that the task itself is often incorrectly specified. For example, while ImageNet is a single image classification tasks, $59 \%$ of ImageNet PIEs codified by humans were identified as multi-object images where multiple labels could be considered reasonable (vs. $39\%$ of non-PIEs). In ImageNet, the over-indexing of incorrectly labelled data and multi-object images in PIE also raises questions about whether the explosion of growth in number of weights in deep neural networks is solving a problem that is better addressed in the data cleaning pipeline. 

\section{Sensitivity of compressed models to distribution shift}\label{sect:distribution_shift}

\begin{figure*}[ht!]
    \centering
\begin{tabular}{cc}
\textbf{ImageNet-A} & \textbf{ImageNet-C} \\
  \includegraphics[width=0.45\columnwidth]{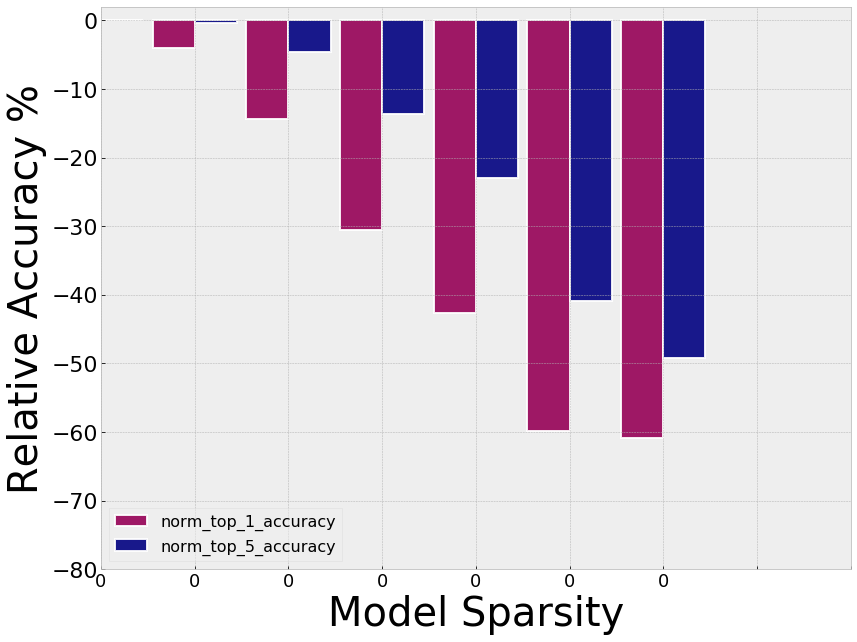}
&		
\includegraphics[width=0.45\columnwidth]{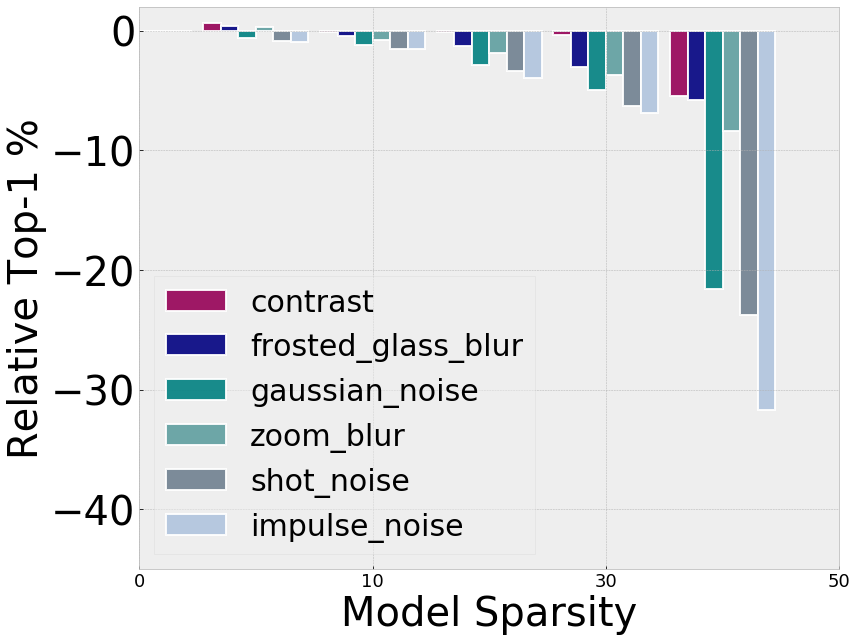}  \\
\end{tabular}
	\caption{
	   High levels of compression amplify sensitivity to distribution shift. \textbf{Left:} Change in top-1 and top-5 recall of a pruned model \textbf{relative} to a non-pruned model on ImageNet-A. \textbf{Right:}  We measure the top-1 test-set performance on a subset of ImageNet-C corruptions of a pruned model relative to the non-pruned model on the same corruption. An extended list of all corruptions considered and top-5 accuracy is included in the supplementary material.
	}
	\label{fig:robustness_imagenet_c_a}
\end{figure*}

Non-compressed models have already been shown to be very brittle to small shifts in the distribution that humans are robust. This can cause unexpected changes in model behavior in the wild that can compromise human welfare \citep{zech2018}. Here, we ask \textit{does compression amplify this brittleness?} Understanding relative differences in robustness helps understand the implications for AI safety of the widespread use of compressed models.

To answer this question, we evaluate the sensitivity of pruned models \emph{relative} to non-pruned models given two open-source benchmarks for robustness: 
\begin{enumerate}
\itemsep0em
\item \textbf{ImageNet-C} \citep{hendrycks2018benchmarking} -- $16$ algorithmically generated corruptions (blur, noise, fog) applied to the ImageNet test-set.
\item \textbf{ImageNet-A} \citep{2019Hendrycks} -- a curated test set of $7,500$ naturally adversarial images designed to produce drastically lower test accuracy.
\end{enumerate}

For each ImageNet-C corruption $q \in Q$, we compare top-1 accuracy of the pruned model evaluated on corruption $q$ normalized by non-pruned model performance on the same corruption. We average across intensities of corruptions as described by \citet{hendrycks2018benchmarking}. If the relative top-1 accuracy was $0$ it would mean that there is no difference in sensitivity to corruptions considered. 

As seen in Fig.~\ref{fig:robustness_imagenet_c_a}, pruning greatly amplifies sensitivity to both ImageNet-C and ImageNet-A relative to non-pruned performance on the same inputs. For ImageNet-C, it is worth noting that relative degradation in performance is remarkably varied across corruptions, with certain corruptions such as \texttt{gaussian}, \texttt{shot noise}, and \texttt{impulse noise} consistently causing far higher relative degradation. At $t=90$, the highest degradation in relative top-1 is \texttt{shot noise} ($-40.11 \%$) and the lowest relative drop is \texttt{brightness} ($-7.73 \%$). Sensitivity to small distribution shifts is amplified at higher levels of sparsity. We include results for all corruptions and the absolute top-1 and top-5 accuracy on each corruption, level of pruning considered in the supplementary material Table. \ref{appendix:ImageNet_A_all_classes}.

The amplified sensitivity of smaller models to distribution shifts and the over-indexing of PIEs on low frequency attributes suggests that much of a models excess capacity is helpful for learning features which aid generalization on atypical or out-of-distribution data points. This builds upon recent work which suggests memorization can benefit generalization properties \citep{2020arXiv200803703F}.

\section{Related work}
The set of model compression techniques is diverse and includes research directions such as reducing the precision or bit size per model weight (quantization) \citep{Jacob_2018, 2014Courbariaux_low_precision_multiplications, Hubara2016_training_neural_networks_low_precision, 2015_gupta}, efforts to start with a network that is more compact with fewer parameters, layers or computations (architecture design) \citep{2017Howard, 2016Squeezenet, kumar17}, student networks with fewer parameters that learn from a larger teacher model (model distillation) \citep{2015hinton} and finally pruning by setting a subset of weights or filters to zero \citep{2017l0_reg, 2016learnedSparsity, Cun90optimalbrain, 1993optimalbrain, Strom97sparseconnection, Hassibi93secondorder, 2017Zhu, 2016abigail, 2017Narang}. In this work, we evaluate the dis-aggregated impact of a subset of pruning and quantization methods.

Despite the widespread use of compression techniques, articulating the trade-offs of compression has overwhelming centered on change to overall accuracy for a given level of compression \citep{Strom97sparseconnection, Cun90optimalbrain, evci2019rigging, 2017Narang, tgale_shooker_2019}. Our work is the first to our knowledge that asks how dis-aggregated measures of model performance at a class and exemplar level are impacted by compression.

In section \ref{sect:distribution_shift}, we also measure sensitivity to two types of distribution shift -- ImageNet-A and ImageNet-C. Recent work by \citep{NIPS2018_7308, Sehwag2019} has considered sensitivity of pruned models to a a different notion of robustness: $l-p$ norm adversarial attacks. In contrast to adversarial robustness which measures the worst-case performance on targeted perturbation, our results provide some understanding of how compressed models perform on subsets of challenging or corrupted natural image examples. \citet{Zhou2019NonvacuousGB} conduct an experiment which shows that networks which are pruned subsequent to training are more sensitive to the corruption of labels at training time.

\section{Discussion and Future Work}\label{sect:discussion_future_work}

The quantization and pruning techniques we evaluate in this paper are already widely used in production systems and integrated with popular deep learning libraries. The popularity and widespread use of these techniques is driven by the severe resource constraints of deploying models to mobile phones or embedded devices \citep{Samala_2018}. Many of the algorithms on your phone are likely pruned or compressed in some way.

Our results suggest that a reliance on top-line metrics such as top-1 or top-5 test-set accuracy hides critical details in the ways that compression impacts model generalization. Caution should be used before deploying compressed models to sensitive domains such as hiring, health care diagnostics, self-driving cars, facial recognition software. For these domains, the introduction of pruning may be at odds with the need to guarantee a certain level of recall or performance for certain subsets of the dataset.

\textbf{Role of Capacity in Deep Neural Networks} A “bigger is better” race in the number of model parameters has gripped the field of machine learning \citep{2016Canziani, 2019arXiv190602243S}. However, the role of additional weights is not well understood. The over-indexing of PIEs on low frequency attributes suggest that non-compressed networks use the majority of capacity to encode a useful representation for these examples. This costly approach to learning an appropriate mapping for a small subset of examples may be better solved in the data pipeline.

\textbf{Auditing and improving compressed models} Our methodology offers one way for humans to better understand the trade-offs incurred by compression and surface challenging examples for human judgement. Identifying harm is the first step in proposing a remedy, and we anticipate our work may spur focus on developing new compression techniques that improve upon the disparate impact we identify and characterize in this work.

\textbf{Limitations} There is substantial ground we were not able to address within the scope of this work. Open questions remain about the implications of these findings for other possible desirable objectives such as fairness.Underserved areas worthy of future consideration include evaluating the impact of compression on additional domains such as language and audio, and leveraging these insights to explicitly optimize for compressed models that \emph{also} minimize the disparate impact on underrepresented data attributes.

\subsubsection*{Acknowledgements} We thank the generosity of our peers for valuable input on earlier versions of this work. In particular, we would like to acknowledge the input of Jonas Kemp, Simon Kornblith, Julius Adebayo, Hugo Larochelle, Dumitru Erhan, Nicolas Papernot, Catherine Olsson, Cliff Young, Martin Wattenberg, Utku Evci, James Wexler, Trevor Gale,  Melissa Fabros, Prajit Ramachandran, Pieter Kindermans, Erich Elsen and Moustapha Cisse. We thank R6 from ICML 2021 for pointing out some improvements to the formulation of the class level metrics. We thank the institutional support and encouragement of Natacha Mainville and Alexander Popper.

\bibliography{main}

\begin{thebibliography}{75}
\providecommand{\natexlab}[1]{#1}
\providecommand{\url}[1]{\texttt{#1}}
\expandafter\ifx\csname urlstyle\endcsname\relax
  \providecommand{\doi}[1]{doi: #1}\else
  \providecommand{\doi}{doi: \begingroup \urlstyle{rm}\Url}\fi

\bibitem[Alvarez et~al.(2016)Alvarez, Prabhavalkar, and Bakhtin]{Alvarez_2016}
Alvarez, R., Prabhavalkar, R., and Bakhtin, A.
\newblock On the efficient representation and execution of deep acoustic
  models.
\newblock \emph{Interspeech 2016}, Sep 2016.
\newblock \doi{10.21437/interspeech.2016-128}.
\newblock URL \url{http://dx.doi.org/10.21437/Interspeech.2016-128}.

\bibitem[Anderson \& Darling(1954)Anderson and Darling]{1954anderson_darling}
Anderson, T.~W. and Darling, D.~A.
\newblock A test of goodness of fit.
\newblock \emph{Journal of the American Statistical Association}, 49\penalty0
  (268):\penalty0 765--769, 1954.
\newblock ISSN 01621459.
\newblock URL \url{http://www.jstor.org/stable/2281537}.

\bibitem[Bartlett \& Wegkamp(2008)Bartlett and Wegkamp]{Bartlett2008}
Bartlett, P.~L. and Wegkamp, M.~H.
\newblock Classification with a reject option using a hinge loss.
\newblock \emph{J. Mach. Learn. Res.}, 9:\penalty0 1823--1840, June 2008.
\newblock ISSN 1532-4435.
\newblock URL \url{http://dl.acm.org/citation.cfm?id=1390681.1442792}.

\bibitem[{Canziani} et~al.(2016){Canziani}, {Paszke}, and
  {Culurciello}]{2016Canziani}
{Canziani}, A., {Paszke}, A., and {Culurciello}, E.
\newblock {An Analysis of Deep Neural Network Models for Practical
  Applications}.
\newblock \emph{arXiv e-prints}, art. arXiv:1605.07678, May 2016.

\bibitem[Caruana(2000)]{Caruana2000}
Caruana, R.
\newblock Case-based explanation for artificial neural nets.
\newblock In Malmgren, H., Borga, M., and Niklasson, L. (eds.),
  \emph{Artificial Neural Networks in Medicine and Biology}, pp.\  303--308,
  London, 2000. Springer London.
\newblock ISBN 978-1-4471-0513-8.

\bibitem[Casey et~al.(2000)Casey, Giedd, and Thomas]{CASEY2000241}
Casey, B., Giedd, J.~N., and Thomas, K.~M.
\newblock Structural and functional brain development and its relation to
  cognitive development.
\newblock \emph{Biological Psychology}, 54\penalty0 (1):\penalty0 241 -- 257,
  2000.
\newblock ISSN 0301-0511.
\newblock \doi{https://doi.org/10.1016/S0301-0511(00)00058-2}.
\newblock URL
  \url{http://www.sciencedirect.com/science/article/pii/S0301051100000582}.

\bibitem[{Chen} et~al.(2016){Chen}, {Emer}, and {Sze}]{7551407Chen}
{Chen}, Y., {Emer}, J., and {Sze}, V.
\newblock Eyeriss: A spatial architecture for energy-efficient dataflow for
  convolutional neural networks.
\newblock In \emph{2016 ACM/IEEE 43rd Annual International Symposium on
  Computer Architecture (ISCA)}, pp.\  367--379, June 2016.
\newblock \doi{10.1109/ISCA.2016.40}.

\bibitem[{Collins} \& {Kohli}(2014){Collins} and {Kohli}]{2014memorybounded}
{Collins}, M.~D. and {Kohli}, P.
\newblock {Memory Bounded Deep Convolutional Networks}.
\newblock \emph{ArXiv e-prints}, December 2014.

\bibitem[Collins \& Kohli(2014)Collins and Kohli]{CollinsK14}
Collins, M.~D. and Kohli, P.
\newblock Memory bounded deep convolutional networks.
\newblock \emph{CoRR}, abs/1412.1442, 2014.
\newblock URL \url{http://arxiv.org/abs/1412.1442}.

\bibitem[Cortes et~al.(2016)Cortes, DeSalvo, and Mohri]{NIPS2016Cortes}
Cortes, C., DeSalvo, G., and Mohri, M.
\newblock Boosting with abstention.
\newblock In Lee, D.~D., Sugiyama, M., Luxburg, U.~V., Guyon, I., and Garnett,
  R. (eds.), \emph{Advances in Neural Information Processing Systems 29}, pp.\
  1660--1668. Curran Associates, Inc., 2016.
\newblock URL
  \url{http://papers.nips.cc/paper/6336-boosting-with-abstention.pdf}.

\bibitem[{Courbariaux} et~al.(2014){Courbariaux}, {Bengio}, and
  {David}]{2014Courbariaux_low_precision_multiplications}
{Courbariaux}, M., {Bengio}, Y., and {David}, J.-P.
\newblock {Training deep neural networks with low precision multiplications}.
\newblock \emph{arXiv e-prints}, art. arXiv:1412.7024, Dec 2014.

\bibitem[Cun et~al.(1990)Cun, Denker, and Solla]{Cun90optimalbrain}
Cun, Y.~L., Denker, J.~S., and Solla, S.~A.
\newblock Optimal brain damage.
\newblock In \emph{Advances in Neural Information Processing Systems}, pp.\
  598--605. Morgan Kaufmann, 1990.

\bibitem[D'Agostino \& Stephens(1986)D'Agostino and Stephens]{1986agostino}
D'Agostino, R.~B. and Stephens, M.~A. (eds.).
\newblock \emph{Goodness-of-fit Techniques}.
\newblock Marcel Dekker, Inc., New York, NY, USA, 1986.
\newblock ISBN 0-824-77487-6.

\bibitem[Deng et~al.(2009)Deng, Dong, Socher, Li, Li, and
  Fei-Fei]{imagenet_cvpr09}
Deng, J., Dong, W., Socher, R., Li, L.-J., Li, K., and Fei-Fei, L.
\newblock {ImageNet: A Large-Scale Hierarchical Image Database}.
\newblock In \emph{CVPR09}, 2009.

\bibitem[Evci et~al.(2019)Evci, Gale, Menick, Castro, and
  Elsen]{evci2019rigging}
Evci, U., Gale, T., Menick, J., Castro, P.~S., and Elsen, E.
\newblock Rigging the lottery: Making all tickets winners, 2019.

\bibitem[{Feldman} \& {Zhang}(2020){Feldman} and {Zhang}]{2020arXiv200803703F}
{Feldman}, V. and {Zhang}, C.
\newblock {What Neural Networks Memorize and Why: Discovering the Long Tail via
  Influence Estimation}.
\newblock \emph{arXiv e-prints}, art. arXiv:2008.03703, August 2020.

\bibitem[Gale et~al.(2019)Gale, Elsen, and Hooker]{tgale_shooker_2019}
Gale, T., Elsen, E., and Hooker, S.
\newblock The state of sparsity in deep neural networks.
\newblock \emph{CoRR}, abs/1902.09574, 2019.
\newblock URL \url{http://arxiv.org/abs/1902.09574}.

\bibitem[Gordon et~al.(2018)Gordon, Eban, Nachum, Chen, Wu, Yang, and
  Choi]{Gordon_2018}
Gordon, A., Eban, E., Nachum, O., Chen, B., Wu, H., Yang, T.-J., and Choi, E.
\newblock Morphnet: Fast \& simple resource-constrained structure learning of
  deep networks.
\newblock \emph{2018 IEEE/CVF Conference on Computer Vision and Pattern
  Recognition}, Jun 2018.
\newblock \doi{10.1109/cvpr.2018.00171}.
\newblock URL \url{http://dx.doi.org/10.1109/CVPR.2018.00171}.

\bibitem[{Guo} et~al.(2017){Guo}, {Pleiss}, {Sun}, and {Weinberger}]{2017Guo}
{Guo}, C., {Pleiss}, G., {Sun}, Y., and {Weinberger}, K.~Q.
\newblock {On Calibration of Modern Neural Networks}.
\newblock \emph{arXiv e-prints}, art. arXiv:1706.04599, Jun 2017.

\bibitem[Guo et~al.(2016)Guo, Yao, and Chen]{Guo2016}
Guo, Y., Yao, A., and Chen, Y.
\newblock Dynamic network surgery for efficient dnns.
\newblock \emph{CoRR}, abs/1608.04493, 2016.
\newblock URL \url{http://arxiv.org/abs/1608.04493}.

\bibitem[Guo et~al.(2018)Guo, Zhang, Zhang, and Chen]{NIPS2018_7308}
Guo, Y., Zhang, C., Zhang, C., and Chen, Y.
\newblock Sparse dnns with improved adversarial robustness.
\newblock In Bengio, S., Wallach, H., Larochelle, H., Grauman, K.,
  Cesa-Bianchi, N., and Garnett, R. (eds.), \emph{Advances in Neural
  Information Processing Systems 31}, pp.\  242--251. Curran Associates, Inc.,
  2018.
\newblock URL
  \url{http://papers.nips.cc/paper/7308-sparse-dnns-with-improved-adversarial-robustness.pdf}.

\bibitem[Gupta et~al.(2015)Gupta, Agrawal, Gopalakrishnan, and
  Narayanan]{2015_gupta}
Gupta, S., Agrawal, A., Gopalakrishnan, K., and Narayanan, P.
\newblock Deep learning with limited numerical precision.
\newblock \emph{CoRR}, abs/1502.02551, 2015.
\newblock URL \url{http://arxiv.org/abs/1502.02551}.

\bibitem[Han et~al.(2015)Han, Pool, Tran, and Dally]{lwac}
Han, S., Pool, J., Tran, J., and Dally, W.~J.
\newblock {L}earning both {W}eights and {C}onnections for {E}fficient {N}eural
  {N}etwork.
\newblock In \emph{{NIPS}}, pp.\  1135--1143, 2015.

\bibitem[Hassibi et~al.(1993{\natexlab{a}})Hassibi, Stork, and
  Com]{Hassibi93secondorder}
Hassibi, B., Stork, D.~G., and Com, S. C.~R.
\newblock Second order derivatives for network pruning: Optimal brain surgeon.
\newblock In \emph{Advances in Neural Information Processing Systems 5}, pp.\
  164--171. Morgan Kaufmann, 1993{\natexlab{a}}.

\bibitem[Hassibi et~al.(1993{\natexlab{b}})Hassibi, Stork, and
  Wolff]{1993optimalbrain}
Hassibi, B., Stork, D.~G., and Wolff, G.~J.
\newblock Optimal brain surgeon and general network pruning.
\newblock In \emph{IEEE International Conference on Neural Networks}, pp.\
  293--299 vol.1, March 1993{\natexlab{b}}.
\newblock \doi{10.1109/ICNN.1993.298572}.

\bibitem[{He} et~al.(2015){He}, {Zhang}, {Ren}, and {Sun}]{He_2015}
{He}, K., {Zhang}, X., {Ren}, S., and {Sun}, J.
\newblock {Deep Residual Learning for Image Recognition}.
\newblock \emph{ArXiv e-prints}, December 2015.

\bibitem[Hendrycks \& Dietterich(2019)Hendrycks and
  Dietterich]{hendrycks2018benchmarking}
Hendrycks, D. and Dietterich, T.
\newblock Benchmarking neural network robustness to common corruptions and
  perturbations.
\newblock In \emph{International Conference on Learning Representations}, 2019.
\newblock URL \url{https://openreview.net/forum?id=HJz6tiCqYm}.

\bibitem[{Hendrycks} et~al.(2019){Hendrycks}, {Zhao}, {Basart}, {Steinhardt},
  and {Song}]{2019Hendrycks}
{Hendrycks}, D., {Zhao}, K., {Basart}, S., {Steinhardt}, J., and {Song}, D.
\newblock {Natural Adversarial Examples}.
\newblock \emph{arXiv e-prints}, art. arXiv:1907.07174, Jul 2019.

\bibitem[{Hinton} et~al.(2015){Hinton}, {Vinyals}, and {Dean}]{2015hinton}
{Hinton}, G., {Vinyals}, O., and {Dean}, J.
\newblock {Distilling the Knowledge in a Neural Network}.
\newblock \emph{arXiv e-prints}, art. arXiv:1503.02531, Mar 2015.

\bibitem[Hooker et~al.(2019)Hooker, Erhan, Kindermans, and Kim]{Hooker2019ABF}
Hooker, S., Erhan, D., Kindermans, P.-J., and Kim, B.
\newblock A benchmark for interpretability methods in deep neural networks.
\newblock In \emph{NeurIPS 2019}, 2019.

\bibitem[{Howard} et~al.(2017){Howard}, {Zhu}, {Chen}, {Kalenichenko}, {Wang},
  {Weyand}, {Andreetto}, and {Adam}]{2017Howard}
{Howard}, A.~G., {Zhu}, M., {Chen}, B., {Kalenichenko}, D., {Wang}, W.,
  {Weyand}, T., {Andreetto}, M., and {Adam}, H.
\newblock {MobileNets: Efficient Convolutional Neural Networks for Mobile
  Vision Applications}.
\newblock \emph{ArXiv e-prints}, April 2017.

\bibitem[Hubara et~al.(2016)Hubara, Courbariaux, Soudry, El{-}Yaniv, and
  Bengio]{Hubara2016_training_neural_networks_low_precision}
Hubara, I., Courbariaux, M., Soudry, D., El{-}Yaniv, R., and Bengio, Y.
\newblock Quantized neural networks: Training neural networks with low
  precision weights and activations.
\newblock \emph{CoRR}, abs/1609.07061, 2016.
\newblock URL \url{http://arxiv.org/abs/1609.07061}.

\bibitem[Huber-Carol et~al.(2002)Huber-Carol, Balakrishnan, Nikulin, and
  Mesbah]{2002huber}
Huber-Carol, C., Balakrishnan, N., Nikulin, M., and Mesbah, M.
\newblock \emph{Goodness-of-Fit Tests and Model Validity}.
\newblock Goodness-of-fit Tests and Model Validity. Birkh{\"a}user Boston,
  2002.
\newblock ISBN 9780817642099.
\newblock URL \url{https://books.google.com/books?id=gUMcv2\_NrhkC}.

\bibitem[{Iandola} et~al.(2016){Iandola}, {Han}, {Moskewicz}, {Ashraf},
  {Dally}, and {Keutzer}]{2016Squeezenet}
{Iandola}, F.~N., {Han}, S., {Moskewicz}, M.~W., {Ashraf}, K., {Dally}, W.~J.,
  and {Keutzer}, K.
\newblock {SqueezeNet: AlexNet-level accuracy with 50x fewer parameters and
  $<$0.5MB model size}.
\newblock \emph{ArXiv e-prints}, February 2016.

\bibitem[Ioffe \& Szegedy(2015)Ioffe and Szegedy]{Ioffe2015}
Ioffe, S. and Szegedy, C.
\newblock Batch normalization: Accelerating deep network training by reducing
  internal covariate shift.
\newblock \emph{CoRR}, abs/1502.03167, 2015.
\newblock URL \url{http://arxiv.org/abs/1502.03167}.

\bibitem[Jacob et~al.(2018)Jacob, Kligys, Chen, Zhu, Tang, Howard, Adam, and
  Kalenichenko]{Jacob_2018}
Jacob, B., Kligys, S., Chen, B., Zhu, M., Tang, M., Howard, A., Adam, H., and
  Kalenichenko, D.
\newblock Quantization and training of neural networks for efficient
  integer-arithmetic-only inference.
\newblock \emph{2018 IEEE/CVF Conference on Computer Vision and Pattern
  Recognition}, Jun 2018.
\newblock \doi{10.1109/cvpr.2018.00286}.
\newblock URL \url{http://dx.doi.org/10.1109/CVPR.2018.00286}.

\bibitem[Kalchbrenner et~al.(2018)Kalchbrenner, Elsen, Simonyan, Noury,
  Casagrande, Lockhart, Stimberg, van~den Oord, Dieleman, and
  Kavukcuoglu]{wavernn}
Kalchbrenner, N., Elsen, E., Simonyan, K., Noury, S., Casagrande, N., Lockhart,
  E., Stimberg, F., van~den Oord, A., Dieleman, S., and Kavukcuoglu, K.
\newblock Efficient {Neural} {Audio} {Synthesis}.
\newblock In \emph{Proceedings of the 35th International Conference on Machine
  Learning, {ICML} 2018, Stockholmsm{\"{a}}ssan, Stockholm, Sweden, July 10-15,
  2018}, pp.\  2415--2424, 2018.

\bibitem[Kendall \& Gal(2017)Kendall and Gal]{NIPSKendall2017}
Kendall, A. and Gal, Y.
\newblock What uncertainties do we need in bayesian deep learning for computer
  vision?
\newblock In Guyon, I., Luxburg, U.~V., Bengio, S., Wallach, H., Fergus, R.,
  Vishwanathan, S., and Garnett, R. (eds.), \emph{Advances in Neural
  Information Processing Systems 30}, pp.\  5574--5584. Curran Associates,
  Inc., 2017.

\bibitem[Kim et~al.(2016)Kim, Khanna, and Koyejo]{NIPS2016_6300}
Kim, B., Khanna, R., and Koyejo, O.~O.
\newblock Examples are not enough, learn to criticize! criticism for
  interpretability.
\newblock In Lee, D.~D., Sugiyama, M., Luxburg, U.~V., Guyon, I., and Garnett,
  R. (eds.), \emph{Advances in Neural Information Processing Systems 29}, pp.\
  2280--2288. Curran Associates, Inc., 2016.

\bibitem[Kolb \& Whishaw(2009)Kolb and Whishaw]{kolb2009fundamentals}
Kolb, B. and Whishaw, I.
\newblock \emph{Fundamentals of Human Neuropsychology}.
\newblock A series of books in psychology. Worth Publishers, 2009.
\newblock ISBN 9780716795865.

\bibitem[Krizhevsky(2012)]{Krizhevsky09learningmultiple}
Krizhevsky, A.
\newblock Learning multiple layers of features from tiny images.
\newblock \emph{University of Toronto}, 05 2012.

\bibitem[Kumar et~al.(2017)Kumar, Goyal, and Varma]{kumar17}
Kumar, A., Goyal, S., and Varma, M.
\newblock Resource-efficient machine learning in 2 {KB} {RAM} for the internet
  of things.
\newblock In Precup, D. and Teh, Y.~W. (eds.), \emph{Proceedings of the 34th
  International Conference on Machine Learning}, volume~70 of \emph{Proceedings
  of Machine Learning Research}, pp.\  1935--1944, International Convention
  Centre, Sydney, Australia, 06--11 Aug 2017. PMLR.
\newblock URL \url{http://proceedings.mlr.press/v70/kumar17a.html}.

\bibitem[Lattner et~al.(2020)Lattner, Amini, Bondhugula, Cohen, Davis, Pienaar,
  Riddle, Shpeisman, Vasilache, and Zinenko]{lattner2020mlir}
Lattner, C., Amini, M., Bondhugula, U., Cohen, A., Davis, A., Pienaar, J.,
  Riddle, R., Shpeisman, T., Vasilache, N., and Zinenko, O.
\newblock Mlir: A compiler infrastructure for the end of moore's law, 2020.

\bibitem[Lee et~al.(2018)Lee, Ajanthan, and Torr]{namhoon2018}
Lee, N., Ajanthan, T., and Torr, P. H.~S.
\newblock {SNIP:} single-shot network pruning based on connection sensitivity.
\newblock \emph{CoRR}, abs/1810.02340, 2018.
\newblock URL \url{http://arxiv.org/abs/1810.02340}.

\bibitem[Leibig et~al.(2017)Leibig, Allken, Ayhan, Berens, and
  Wahl]{Leibig2017}
Leibig, C., Allken, V., Ayhan, M.~S., Berens, P., and Wahl, S.
\newblock Leveraging uncertainty information from deep neural networks for
  disease detection.
\newblock \emph{Scientific Reports}, 7, 12 2017.
\newblock \doi{10.1038/s41598-017-17876-z}.

\bibitem[Li et~al.(2020)Li, Wallace, Shen, Lin, Keutzer, Klein, and
  Gonzalez]{li2020train}
Li, Z., Wallace, E., Shen, S., Lin, K., Keutzer, K., Klein, D., and Gonzalez,
  J.~E.
\newblock Train large, then compress: Rethinking model size for efficient
  training and inference of transformers, 2020.

\bibitem[Liu et~al.(2015)Liu, Luo, Wang, and Tang]{liu2015faceattributes}
Liu, Z., Luo, P., Wang, X., and Tang, X.
\newblock Deep learning face attributes in the wild.
\newblock In \emph{Proceedings of International Conference on Computer Vision
  (ICCV)}, December 2015.

\bibitem[{Liu} et~al.(2017){Liu}, {Li}, {Shen}, {Huang}, {Yan}, and
  {Zhang}]{2017Liu}
{Liu}, Z., {Li}, J., {Shen}, Z., {Huang}, G., {Yan}, S., and {Zhang}, C.
\newblock {Learning Efficient Convolutional Networks through Network Slimming}.
\newblock \emph{ArXiv e-prints}, August 2017.

\bibitem[{Louizos} et~al.(2017){Louizos}, {Welling}, and {Kingma}]{2017l0_reg}
{Louizos}, C., {Welling}, M., and {Kingma}, D.~P.
\newblock {Learning Sparse Neural Networks through $L\_0$ Regularization}.
\newblock \emph{ArXiv e-prints}, December 2017.

\bibitem[{Micikevicius} et~al.(2017){Micikevicius}, {Narang}, {Alben},
  {Diamos}, {Elsen}, {Garcia}, {Ginsburg}, {Houston}, {Kuchaiev}, {Venkatesh},
  and {Wu}]{2017Micikevicius}
{Micikevicius}, P., {Narang}, S., {Alben}, J., {Diamos}, G., {Elsen}, E.,
  {Garcia}, D., {Ginsburg}, B., {Houston}, M., {Kuchaiev}, O., {Venkatesh}, G.,
  and {Wu}, H.
\newblock {Mixed Precision Training}.
\newblock \emph{arXiv e-prints}, art. arXiv:1710.03740, October 2017.

\bibitem[{Narang} et~al.(2017){Narang}, {Elsen}, {Diamos}, and
  {Sengupta}]{2017Narang}
{Narang}, S., {Elsen}, E., {Diamos}, G., and {Sengupta}, S.
\newblock {Exploring Sparsity in Recurrent Neural Networks}.
\newblock \emph{arXiv e-prints}, art. arXiv:1704.05119, Apr 2017.

\bibitem[Nowlan \& Hinton(1992)Nowlan and Hinton]{1992_nowlan_hinton}
Nowlan, S.~J. and Hinton, G.~E.
\newblock Simplifying neural networks by soft weight-sharing.
\newblock \emph{Neural Computation}, 4\penalty0 (4):\penalty0 473--493, 1992.
\newblock \doi{10.1162/neco.1992.4.4.473}.
\newblock URL \url{https://doi.org/10.1162/neco.1992.4.4.473}.

\bibitem[Rakic et~al.(1994)Rakic, Bourgeois, and Goldman-Rakic]{RAKI1994}
Rakic, P., Bourgeois, J.-P., and Goldman-Rakic, P.~S.
\newblock Synaptic development of the cerebral cortex: implications for
  learning, memory, and mental illness.
\newblock In Pelt, J.~V., Corner, M., Uylings, H., and Silva, F. L.~D. (eds.),
  \emph{The Self-Organizing Brain: From Growth Cones to Functional Networks},
  volume 102 of \emph{Progress in Brain Research}, pp.\  227 -- 243. Elsevier,
  1994.
\newblock \doi{https://doi.org/10.1016/S0079-6123(08)60543-9}.
\newblock URL
  \url{http://www.sciencedirect.com/science/article/pii/S0079612308605439}.

\bibitem[{Reagen} et~al.(2016){Reagen}, {Whatmough}, {Adolf}, {Rama}, {Lee},
  {Lee}, {Hernández-Lobato}, {Wei}, and {Brooks}]{Reagen_7551399}
{Reagen}, B., {Whatmough}, P., {Adolf}, R., {Rama}, S., {Lee}, H., {Lee},
  S.~K., {Hernández-Lobato}, J.~M., {Wei}, G., and {Brooks}, D.
\newblock Minerva: Enabling low-power, highly-accurate deep neural network
  accelerators.
\newblock In \emph{2016 ACM/IEEE 43rd Annual International Symposium on
  Computer Architecture (ISCA)}, pp.\  267--278, June 2016.
\newblock \doi{10.1109/ISCA.2016.32}.

\bibitem[Samala et~al.(2018)Samala, Chan, Hadjiiski, Helvie, Richter, and
  Cha]{Samala_2018}
Samala, R.~K., Chan, H.-P., Hadjiiski, L.~M., Helvie, M.~A., Richter, C., and
  Cha, K.
\newblock Evolutionary pruning of transfer learned deep convolutional neural
  network for breast cancer diagnosis in digital breast tomosynthesis.
\newblock \emph{Physics in Medicine {\&} Biology}, 63\penalty0 (9):\penalty0
  095005, may 2018.
\newblock \doi{10.1088/1361-6560/aabb5b}.

\bibitem[{See} et~al.(2016){See}, {Luong}, and {Manning}]{2016abigail}
{See}, A., {Luong}, M.-T., and {Manning}, C.~D.
\newblock {Compression of Neural Machine Translation Models via Pruning}.
\newblock \emph{arXiv e-prints}, art. arXiv:1606.09274, Jun 2016.

\bibitem[Sehwag et~al.(2019)Sehwag, Wang, Mittal, and Jana]{Sehwag2019}
Sehwag, V., Wang, S., Mittal, P., and Jana, S.
\newblock Towards compact and robust deep neural networks.
\newblock \emph{CoRR}, abs/1906.06110, 2019.
\newblock URL \url{http://arxiv.org/abs/1906.06110}.

\bibitem[Sowell et~al.(2004)Sowell, Thompson, Leonard, Welcome, Kan, and
  Toga]{Sowell8223}
Sowell, E.~R., Thompson, P.~M., Leonard, C.~M., Welcome, S.~E., Kan, E., and
  Toga, A.~W.
\newblock Longitudinal mapping of cortical thickness and brain growth in normal
  children.
\newblock \emph{Journal of Neuroscience}, 24\penalty0 (38):\penalty0
  8223--8231, 2004.
\newblock \doi{10.1523/JNEUROSCI.1798-04.2004}.
\newblock URL \url{https://www.jneurosci.org/content/24/38/8223}.

\bibitem[{Strubell} et~al.(2019){Strubell}, {Ganesh}, and
  {McCallum}]{2019arXiv190602243S}
{Strubell}, E., {Ganesh}, A., and {McCallum}, A.
\newblock {Energy and Policy Considerations for Deep Learning in NLP}.
\newblock \emph{arXiv e-prints}, art. arXiv:1906.02243, June 2019.

\bibitem[Ström(1997)]{Strom97sparseconnection}
Ström, N.
\newblock Sparse connection and pruning in large dynamic artificial neural
  networks, 1997.

\bibitem[Tessera et~al.(2021)Tessera, Hooker, and Rosman]{tessera2021}
Tessera, K., Hooker, S., and Rosman, B.
\newblock Keep the gradients flowing: Using gradient flow to study sparse
  network optimization.
\newblock \emph{CoRR}, abs/2102.01670, 2021.
\newblock URL \url{https://arxiv.org/abs/2102.01670}.

\bibitem[Theis et~al.(2018)Theis, Korshunova, Tejani, and
  Husz{\'{a}}r]{fisher-pruning}
Theis, L., Korshunova, I., Tejani, A., and Husz{\'{a}}r, F.
\newblock Faster gaze prediction with dense networks and {F}isher pruning.
\newblock \emph{CoRR}, abs/1801.05787, 2018.
\newblock URL \url{http://arxiv.org/abs/1801.05787}.

\bibitem[Ullrich et~al.(2017)Ullrich, Meeds, and Welling]{sws}
Ullrich, K., Meeds, E., and Welling, M.
\newblock Soft {W}eight-{S}haring for {N}eural {N}etwork {C}ompression.
\newblock \emph{CoRR}, abs/1702.04008, 2017.

\bibitem[Valin \& Skoglund(2018)Valin and Skoglund]{lpcnet}
Valin, J. and Skoglund, J.
\newblock Lpcnet: {I}mproving {N}eural {S}peech {S}ynthesis {T}hrough {L}inear
  {P}rediction.
\newblock \emph{CoRR}, abs/1810.11846, 2018.
\newblock URL \url{http://arxiv.org/abs/1810.11846}.

\bibitem[Vanhoucke et~al.(2011)Vanhoucke, Senior, and Mao]{Vanhoucke_2011}
Vanhoucke, V., Senior, A., and Mao, M.~Z.
\newblock Improving the speed of neural networks on cpus.
\newblock In \emph{Deep Learning and Unsupervised Feature Learning Workshop,
  NIPS 2011}, 2011.

\bibitem[Weigend et~al.(1991)Weigend, Rumelhart, and
  Huberman]{NIPS1990_Andreas_weight_elimination}
Weigend, A.~S., Rumelhart, D.~E., and Huberman, B.~A.
\newblock Generalization by weight-elimination with application to forecasting.
\newblock In Lippmann, R.~P., Moody, J.~E., and Touretzky, D.~S. (eds.),
  \emph{Advances in Neural Information Processing Systems 3}, pp.\  875--882.
  Morgan-Kaufmann, 1991.

\bibitem[Welch(1947)]{1947welch}
Welch, B.~L.
\newblock The generalization of `{S}tudent's' problem when several different
  population variances are involved.
\newblock \emph{Biometrika}, 34:\penalty0 28--35, 1947.
\newblock ISSN 0006-3444.
\newblock \doi{10.2307/2332510}.
\newblock URL \url{https://doi.org/10.2307/2332510}.

\bibitem[{Wen} et~al.(2016){Wen}, {Wu}, {Wang}, {Chen}, and
  {Li}]{2016learnedSparsity}
{Wen}, W., {Wu}, C., {Wang}, Y., {Chen}, Y., and {Li}, H.
\newblock {Learning Structured Sparsity in Deep Neural Networks}.
\newblock \emph{ArXiv e-prints}, August 2016.

\bibitem[Zagoruyko \& Komodakis(2016)Zagoruyko and Komodakis]{Zagoruyko2016}
Zagoruyko, S. and Komodakis, N.
\newblock Wide residual networks.
\newblock \emph{CoRR}, abs/1605.07146, 2016.
\newblock URL \url{http://arxiv.org/abs/1605.07146}.

\bibitem[Zech et~al.(2018)Zech, Badgeley, Liu, Costa, Titano, and
  Oermann]{zech2018}
Zech, J.~R., Badgeley, M.~A., Liu, M., Costa, A.~B., Titano, J.~J., and
  Oermann, E.~K.
\newblock Variable generalization performance of a deep learning model to
  detect pneumonia in chest radiographs: A cross-sectional study.
\newblock \emph{PLOS Medicine}, 15\penalty0 (11):\penalty0 1--17, 11 2018.
\newblock \doi{10.1371/journal.pmed.1002683}.
\newblock URL \url{https://doi.org/10.1371/journal.pmed.1002683}.

\bibitem[Zhang(1992)]{ZHANG1992}
Zhang, J.
\newblock Selecting typical instances in instance-based learning.
\newblock In Sleeman, D. and Edwards, P. (eds.), \emph{Machine Learning
  Proceedings 1992}, pp.\  470 -- 479. Morgan Kaufmann, San Francisco (CA),
  1992.
\newblock ISBN 978-1-55860-247-2.
\newblock \doi{https://doi.org/10.1016/B978-1-55860-247-2.50066-8}.
\newblock URL
  \url{http://www.sciencedirect.com/science/article/pii/B9781558602472500668}.

\bibitem[Zhou et~al.(2019)Zhou, Veitch, Austern, Adams, and
  Orbanz]{Zhou2019NonvacuousGB}
Zhou, W., Veitch, V., Austern, M., Adams, R.~P., and Orbanz, P.
\newblock Non-vacuous generalization bounds at the imagenet scale: a
  pac-bayesian compression approach.
\newblock In \emph{ICLR}, 2019.

\bibitem[{Zhu} \& {Gupta}(2017){Zhu} and {Gupta}]{2017Zhu}
{Zhu}, M. and {Gupta}, S.
\newblock {To prune, or not to prune: exploring the efficacy of pruning for
  model compression}.
\newblock \emph{ArXiv e-prints}, October 2017.

\bibitem[Zhu \& Gupta(2017{\natexlab{a}})Zhu and Gupta]{Gupta2017}
Zhu, M. and Gupta, S.
\newblock To prune, or not to prune: exploring the efficacy of pruning for
  model compression.
\newblock \emph{CoRR}, abs/1710.01878, 2017{\natexlab{a}}.
\newblock URL \url{http://arxiv.org/abs/1710.01878}.

\bibitem[Zhu \& Gupta(2017{\natexlab{b}})Zhu and Gupta]{zhu2017prune}
Zhu, M. and Gupta, S.
\newblock To prune, or not to prune: exploring the efficacy of pruning for
  model compression, 2017{\natexlab{b}}.

\end{thebibliography}
\bibliographystyle{neurips_2020}

\clearpage
\appendix

\section*{Appendix}

\begin{table*}[p]
  \centering
\begin{tabular}{cccccc}
\toprule
 \multicolumn{6}{l}{\textbf{ImageNet Robustness to ImageNet-C Corruptions (By Level of Pruning)}} \\
\textbf{Pruning Fraction} & \textbf{Corruption Type} & \textbf{Top-1 }  & \textbf{Top-5} & \textbf{Top-1 Norm} &  \textbf{Top-5 Norm} \\
\midrule
  0.0 &          brightness &           69.49 &           88.98 &                 0.00 &                 0.00 \\
             0.7 &          brightness &           67.50 &           87.86 &                -2.87 &                -1.25 \\
             0.9 &          brightness &           64.12 &           85.63 &                -7.74 &                -3.77 \\
         \midrule
                       
             0.0 &            contrast &           42.30 &           61.80 &                 0.00 &                 0.00 \\
             0.7 &            contrast &           41.34 &           61.58 &                -2.26 &                -0.36 \\
             0.9 &            contrast &           38.04 &           58.43 &               -10.06 &                -5.45 \\

          \midrule
      0.0 &        defocus\_blur &           49.77 &           72.45 &                 0.00 &                 0.00 \\
             0.7 &        defocus\_blur &           47.49 &           70.69 &                -4.58 &                -2.43 \\
             0.9 &        defocus\_blur &           44.69 &           68.26 &               -10.22 &                -5.79 \\

      \midrule
              0.0 &             elastic &           57.09 &           76.71 &                 0.00 &                 0.00 \\
             0.7 &             elastic &           55.09 &           75.29 &                -3.51 &                -1.85 \\
             0.9 &             elastic &           52.81 &           73.62 &                -7.50 &                -4.02 \\

            \midrule
             0.0 &                 fog &           56.21 &           79.25 &                 0.00 &                 0.00 \\
             0.7 &                 fog &           54.46 &           78.25 &                -3.12 &                -1.25 \\
             0.9 &                 fog &           50.36 &           75.10 &               -10.41 &                -5.23 \\

                \midrule
                      0.0 &  frosted\_glass\_blur &           40.89 &           60.51 &                 0.00 &                 0.00 \\
             0.7 &  frosted\_glass\_blur &           38.75 &           58.68 &                -5.23 &                -3.03 \\
             0.9 &  frosted\_glass\_blur &           36.87 &           57.02 &                -9.83 &                -5.78 \\

     \midrule
     0.0 &      gaussian\_noise &           45.43 &           65.67 &                 0.00 &                 0.00 \\
             0.7 &      gaussian\_noise &           42.01 &           62.40 &                -7.53 &                -4.98 \\
             0.9 &      gaussian\_noise &           32.88 &           51.49 &               -27.64 &               -21.59 \\

      \midrule
    0.0 &       impulse\_noise &           42.23 &           63.16 &                 0.00 &                 0.00 \\
             0.7 &       impulse\_noise &           37.91 &           58.82 &               -10.24 &                -6.87 \\
             0.9 &       impulse\_noise &           25.29 &           43.13 &               -40.12 &               -31.70 \\

  \midrule
         0.0 &    jpeg\_compression &           65.75 &           86.25 &                 0.00 &                 0.00 \\
             0.7 &    jpeg\_compression &           63.47 &           84.81 &                -3.47 &                -1.68 \\
             0.9 &    jpeg\_compression &           60.57 &           82.77 &                -7.88 &                -4.04 \\

          \midrule
              0.0 &            pixelate &           57.34 &           78.05 &                 0.00 &                 0.00 \\
             0.7 &            pixelate &           54.93 &           76.17 &                -4.21 &                -2.41 \\
             0.9 &            pixelate &           51.31 &           72.98 &               -10.51 &                -6.50 \\
          
          \midrule
          
             0.0 &          shot\_noise &           43.82 &           64.06 &                 0.00 &                 0.00 \\
             0.7 &          shot\_noise &           39.88 &           60.04 &                -8.99 &                -6.28 \\
             0.9 &          shot\_noise &           30.80 &           48.86 &               -29.71 &               -23.72 \\
             \midrule
             0.0 &           zoom\_blur &           37.16 &           58.90 &                 0.00 &                 0.00 \\
             0.7 &           zoom\_blur &           34.60 &           56.68 &                -6.89 &                -3.76 \\
             0.9 &           zoom\_blur &           31.78 &           53.97 &               -14.47 &                -8.37 \\
\bottomrule
\end{tabular}
 \caption{Pruned models are more sensitive to image corruptions that are meaningless to a human. We measure the average top-1 and top-5 test set accuracy of models trained to varying levels of pruning on the ImageNet-C test-set (the models were trained on uncorrupted ImageNet). For each corruption type, we report the average accuracy of $50$ trained models relative to the baseline models across all $5$ levels of pruning.}\label{appendix:ImageNet_C_all_classes} 
\end{table*}

\section{Pruning and quantization techniques considered}

\paragraph{Magnitude pruning} There are various pruning methodologies that use the absolute value of weights to rank their importance and remove weights that are below a user-specified threshold \citep{CollinsK14,Guo2016, Gupta2017}. These works largely differ in whether the weights are removed permanently or can ``recover" by still receiving subsequent gradient updates. This would allow certain weights to become non-zero again if pruned incorrectly. While magnitude pruning is often used as a criteria to remove individual weights, it can be adapted to remove entire neurons or filters by extending the ranking criteria to a set of weights and setting the threshold appropriately \citep{Gordon_2018}.

In this work, we use the magnitude pruning methodology as proposed by \cite{Gupta2017}. It has been shown to outperform more sophisticated Bayesian pruning methods and is considered state-of-the-art across both computer vision and language models \citep{tgale_shooker_2019}. The choice of magnitude pruning also allowed us to specify and precisely vary the final model sparsity for purposes of our analysis, unlike regularizer approaches that allow the optimization process itself to determine the final level of sparsity \citep{2017Liu, 2017l0_reg, 2014memorybounded, 2016learnedSparsity, NIPS1990_Andreas_weight_elimination, 1992_nowlan_hinton}.

\paragraph{Quantization}
All networks were trained with 32-bit floating point weights and quantized post-training. This means there is no additional gradient updates to the weights post-quantization. In this work, we evaluate three different quantization methods. The first type replaces the weights with 16-bit floating point weights \citep{2017Micikevicius}. The second type  quantizes all weights to 8-bit integer values \citep{Alvarez_2016}. The third type uses the first 100 training examples of each dataset as representative examples for the fixed-point only models. We chose to benchmark these quantization methods in part because each has open source code available. We use TensorFlow Lite with MLIR \citep{lattner2020mlir}.

\begin{table*}[t]
\begin{center}
\begin{tabular}{ccccc}
\toprule
\multicolumn{3}{c}{\textbf{CelebA}} \\
\midrule
\textbf{Fraction Pruned} & \textbf{Top 1}
& \textbf{\# PIEs} \\
\midrule
 0  & 94.73  &  -  \\
0.3 & 94.75 &  555 \\
0.5 & 94.81  &  638 \\ 
0.7 & 94.44 &  990 \\
0.9 & 94.07 &   3229\\
0.95 & 93.39 &  5057\\
0.99 & 90.98 &  8754\\
\midrule
\textbf{Quantization} & \textbf{Top 1}
& \textbf{\# PIEs} \\
hybrid int8 & 94.65 & 404 \\
fixed-point int8 & 94.65 & 414 \\
\bottomrule
\end{tabular}
 \end{center}
 \caption{CelebA top-1 accuracy at all levels of pruning, averaged over runs. The task we consider for CelebA is a binary classification method. We consider exemplar level divergence and classify Pruning Identified Exemplars as the examples where the modal label differs between a population of $30$ compressed and non-compressed models.  Note that the CelebA task is a binary classification task to predict whether the celebrity is blond or non-blond. Thus, there are only two classes.}
 \label{table:appendix_celeba_summary} 
\end{table*}

\section{Pruning Protocol} 

We prune over the course of training to obtain a target end pruning level $t \in \{0.0, 0.1, 0.3, 0.5, 0.7, 0.9\}$.  Removed weights continue to receive gradient updates after being pruned. These hyperparameter choices were based upon a limited grid search which suggested that these particular settings minimized degradation to test-set accuracy across all pruning levels. We note that for CelebA we were able to still converge to a comparable final performance at much higher levels of pruning $t \in \{0.95, 0.99\}$. We include these results, and note that the tolerance for extremely high levels of pruning may be related the relative difficulty of the task. Unlike CIFAR-10 and ImageNet which involve more than $2$ classes ($10$ and $1000$ respectively), CelebA is a binary classification problem. Here, the task is predicting hair color $Y = \{\text{blonde, dark haired}\} $. 

Quantization techniques are applied post-training - the weights are not re-calibrated after quantizing. Figure~\ref{fig:modeldist_all} shows the distributions of model accuracy across model populations for the pruned and quantized models for ImageNet, CIFAR-10 and CelebA. Table. \ref{table:appendix_celeba_summary} and Table. \ref{table:appendix_cifar_imagenet_summary} include top-line metrics for all compression methods considered.

\begin{table*}[t]
\begin{center}
\vskip 0.2in 
\begin{tabular}{cccc}
\toprule
\multicolumn{4}{c}{\textbf{ImageNet}} \\
\midrule
\textbf{Fraction Pruned} & \textbf{Top 1}
& \textbf{\# Signif classes} 
& \textbf{\# PIEs} \\
\midrule
  0    & 76.68 & - & - \\
  30 & 76.46  & 68 & 1,819 \\
  50 & 75.87  & 170 & 2,193 \\
  70 & 75.02  & 372 & 3,073 \\
  90 & 72.60  & 637 & 5,136 \\
 \midrule
 \textbf{Quantization} &   &  &  \\
float16 & 76.65   & 58 & 2019 \\
dynamic range int8 & 76.10  & 144 & 2193  \\
fixed-point int8 & 76.46  & 119 & 2093  \\
\midrule
\multicolumn{4}{c}{\textbf{CIFAR-10}} \\
\midrule
\textbf{Fraction Pruned} & \textbf{Top 1}
& \textbf{\# Signif classes} 
& \textbf{\# PIEs} \\
\midrule
 0  & 94.53 & - &  -  \\
30 & 94.47 & 1 & 114 \\
50 & 94.39 & 1 & 144 \\
70 & 94.30 & 0 & 137 \\
90 & 94.14 & 2 & 216 \\
\bottomrule
\end{tabular}
 \end{center}
 \caption{CIFAR-10 and ImageNet top-1 accuracy at all levels of pruning, averaged over $30$ runs. Top-5 accuracy for CIFAR-10 was $99.8\%$ for all levels of pruning. The third column is the number of classes significantly impacted by pruning.}
 \label{table:appendix_cifar_imagenet_summary} 
\end{table*}

\begin{table*}[t]
\centering
\begin{tabular}{ccccccc}
\toprule
& \multicolumn{3}{c}{\textbf{Top-1 accuracy}} & \multicolumn{3}{c}{\textbf{Top-5 accuracy}} \\
\midrule
\multicolumn{7}{c}{\textbf{ImageNet}} \\
\midrule
\textbf{Fraction Pruned} & \textbf{Non-PIEs} &  \textbf{PIEs} & \textbf{All} &  \textbf{Non-PIEs} &   \textbf{PIEs} & \textbf{All} \\
\midrule
           10.0 &          79.34 &  26.14 &   76.75  &          94.89 &  68.52 &    93.35 \\
           30.0 &          79.23 &  26.21 &    76.75 &          95.04 &  69.30 &    93.35 \\
           50.0 &          79.54 &  28.74 &    76.75 &          94.89 &  71.47 &    93.35 \\
           70.0 &          80.16 &  32.06 &    76.75 &          94.99 &  74.74 &    93.35 \\
           90.0 &          81.20 &  39.81 &  76.75 &          95.11 &  78.90 &  93.35 \\
\midrule         
\multicolumn{7}{c}{\textbf{CIFAR-10}} \\
\midrule    
\textbf{Fraction Pruned} & \textbf{Non-PIEs} &  \textbf{PIEs} & \textbf{All} &  \textbf{Non-PIEs} &   \textbf{PIEs} & \textbf{All} \\
\midrule
         10.0 &   95.11 &    43.23 &       94.89 &   99.91 & 95.30 &      99.91 \\
                      30.0 &           95.40 &    40.61 &       94.89 &           99.92 &      92.83 &     99.91 \\
                      50.0 &           95.45 &   40.42 &        94.89 &            99.93 &     93.53 &     99.91 \\
                      70.0 &         95.56 &        43.64 &     94.89 &           99.94 &      95.95 &     99.91 \\
                      90.0 &             95.60 &   50.71 &      94.89 &           99.92 &      96.67 &     99.91 \\
\multicolumn{7}{c}{\textbf{CelebA}} \\
\midrule    
\textbf{Fraction Pruned} & \textbf{Non-PIEs} &  \textbf{PIEs} & \textbf{All} &  \textbf{Non-PIEs} &   \textbf{PIEs} & \textbf{All} \\
\midrule
   30.0 &          94.76 &                        49.82 &                                  94.76 & - & - & -  \\
           50.0 &          94.78 &                       50.55 &                                  94.78  & - & - & -   \\
           70.0 &          94.54 &                         52.61 &                                  94.54  & - & - & -  \\
           90.0 &          94.10 &                         50.41 &                                  94.10  & - & - & -  \\
           95.0 &          93.40 &                        45.57 &                                  93.40  & - & - & -  \\
           99.0 &          90.97 &                       39.84 &                                  90.97  & - & - & -  \\
\bottomrule
\end{tabular}
\caption{A comparison of non-compressed model performance on Pruning Identified Exemplars (PIE) relative to a random sample drawn independently from the test-set and a sample excluding PIEs (non-PIEs). Inference on the non-PIE sample improves test-set top-1 accuracy relative to the baseline for ImageNet and Cifar-10. Evaluation on PIE images alone yields substantially lower top-1 accuracy. Note that CelebA top-5 is not included as it is a binary classification problem.}
\label{table:appendix_test_set_accuracy} 
\end{table*}

\begin{figure*}[t]
\centering
 \includegraphics[width=.9\textwidth]{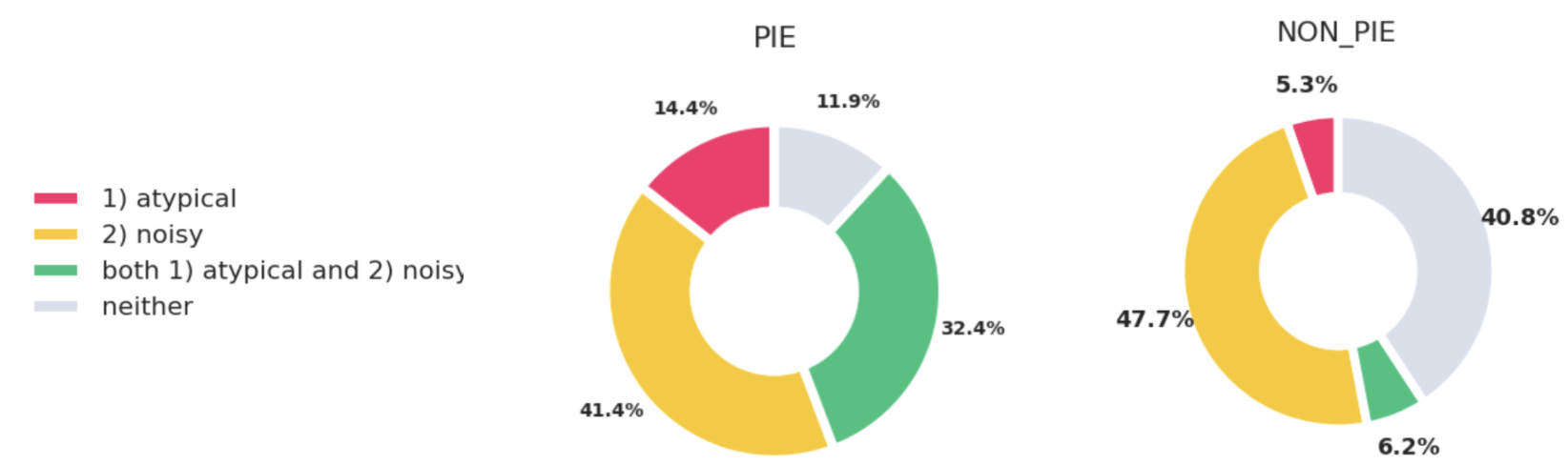}
  \caption{A pie chart of the codified attributes of a sample of pruning identified examplars (PIEs) and non-PIE images. The human study shows that PIEs over-index on both \textbf{noisy} exemplars with partial or corrupt information (corrupted images, incorrect labels, multi-object images) and/or \textbf{atypical} or challenging images (abstract representation, fine grained classification).}
\label{fig:pie_pie}
\vskip -0.1in 
\end{figure*}

\section{Human study}

\begin{table*}
    \centering
    \begin{tabular}{ccc}
    \includegraphics[width=0.3\textwidth]{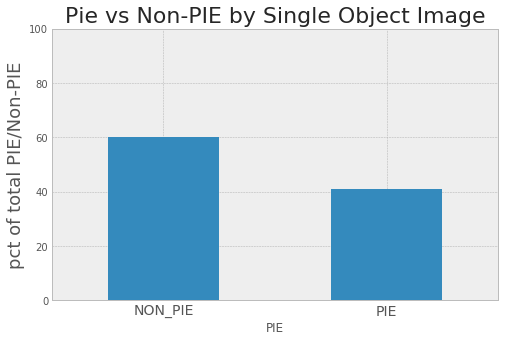} &
    \includegraphics[width=0.3\textwidth]{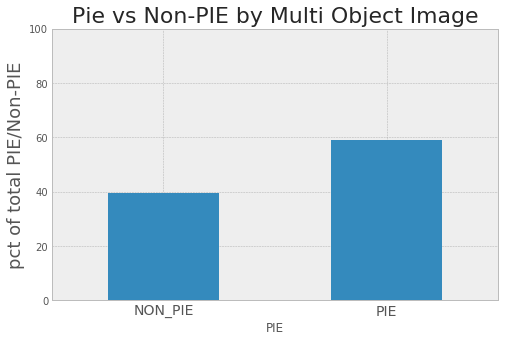} &
    \includegraphics[width=0.3\textwidth]{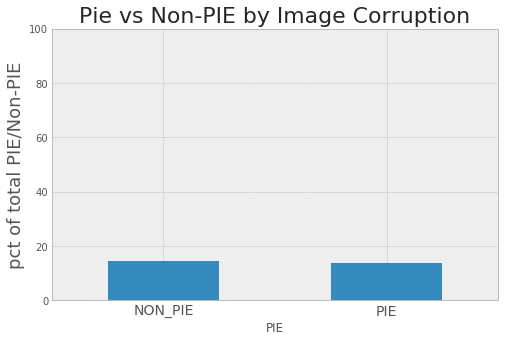} \\ 
    \includegraphics[width=0.3\textwidth]{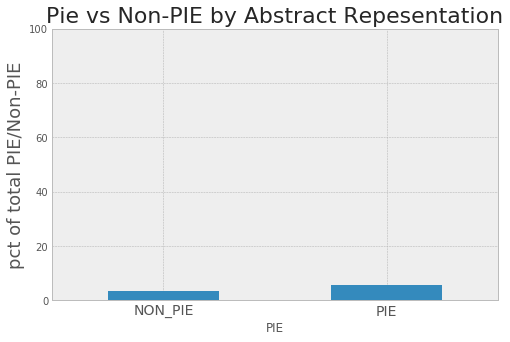} &
    \includegraphics[width=0.3\textwidth]{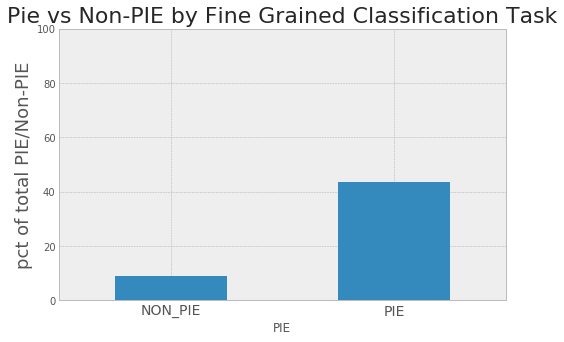} &
    \includegraphics[width=0.3\textwidth]{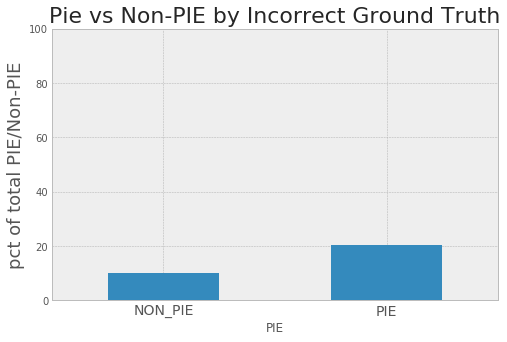}
    \end{tabular}
    \caption{PIE vs non-PIE relative representation for different attributes. These attributes were codified in a human study involving 85 individuals inspecting a balanced random sample of PIE and non-PIE.  The classification as PIE or non-PIE was not known or available to the human.}
    \label{fig:human_study}
\end{table*}

We conducted a human study (involving $85$ volunteers) to label a random sample of $1230$ PIE and non-PIE ImageNet images. Humans in the study were shown a balanced sample of PIE and non-PIE images that were selected at random and shuffled. The classification as PIE or non-PIE was not known or available to the human. Participants answered the following questions for each image that was presented:
\begin{itemize}
 \item \textit{Does label 1 accurately label an object in the image? (0/1)}
 \item \textit{Does this image depict a single object? (0/1)}
 \item  \textit{Would you consider labels 1, 2 and 3 to be semantically very close to each other? (does this image require fine grained classification) (0/1)}
 \item  \textit{Do you consider the object in the image to be a typical exemplar for the class indicated by label 1? (0/1)}
 \item  \textit{Is the image quality corrupted (some common image corruptions -- overlaid text, brightness, contrast, filter, defocus blur, fog,
 jpeg compression, pixelate, shot noise, zoom blur, black and white vs. rgb)? (0/1)}
 \item \textit{Is the object in the image an abstract representation of the class indicated by label 1? [[an abstract representation is an object in an abstract form, such as a painting, drawing or rendering using a different material.]] (0/1)}
\end{itemize}

We find that PIEs heavily over-index relative to non-PIEs on both \textbf{noisy} examples with corrupted information (incorrect ground truth label, multiple objects, image corruption) and \textbf{atypical} or challenging examples (fine-grained classification task, abstract representation). We include the per attribute relative representation of PIE vs. Non-PIE for the study (in Figure. \ref{fig:human_study}). 

\begin{figure*}[hb!]
\includegraphics[width=1\textwidth]{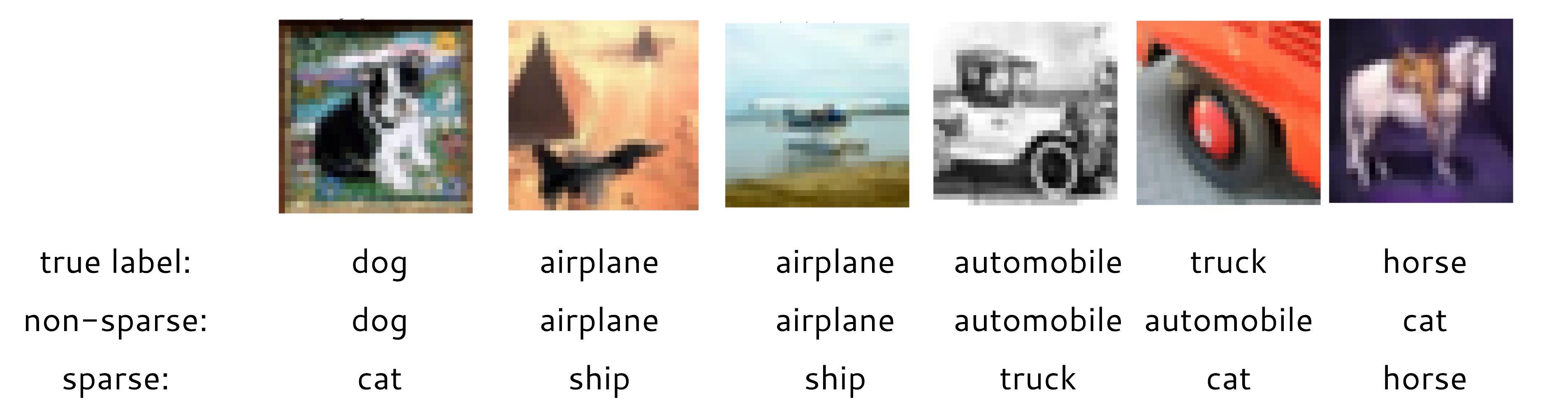} \caption{Visualization of Pruning Identified Exemplars from the CIFAR-10 dataset. This subset of impacted images is identified by considering a set of $30$ non-pruned wide ResNet models and $30$ models trained to $30 \%$ pruning. Below each image are three labels: 1) true label, 2) the modal (most frequent) prediction from the set of non-pruned models, 3) the modal prediction from the set of pruned models.}
\label{fig:corrupt_abstract_all_cifar}
\vskip -0.1in 
\end{figure*}

\section{Benchmarks to evaluate robustness}

\begin{figure*}[ht!] 
\begin{center}
  \includegraphics[width=0.4\textwidth]{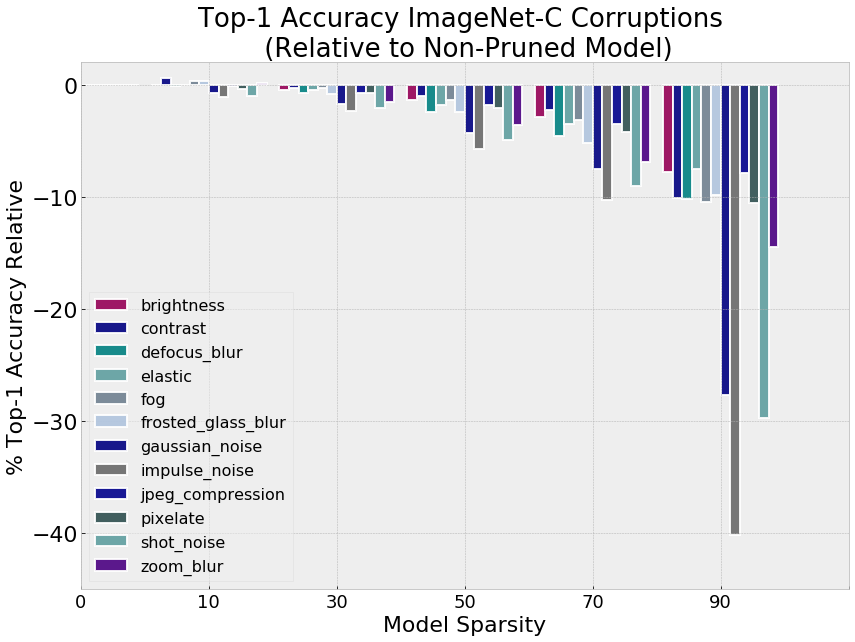}
    \includegraphics[width=0.4\textwidth]{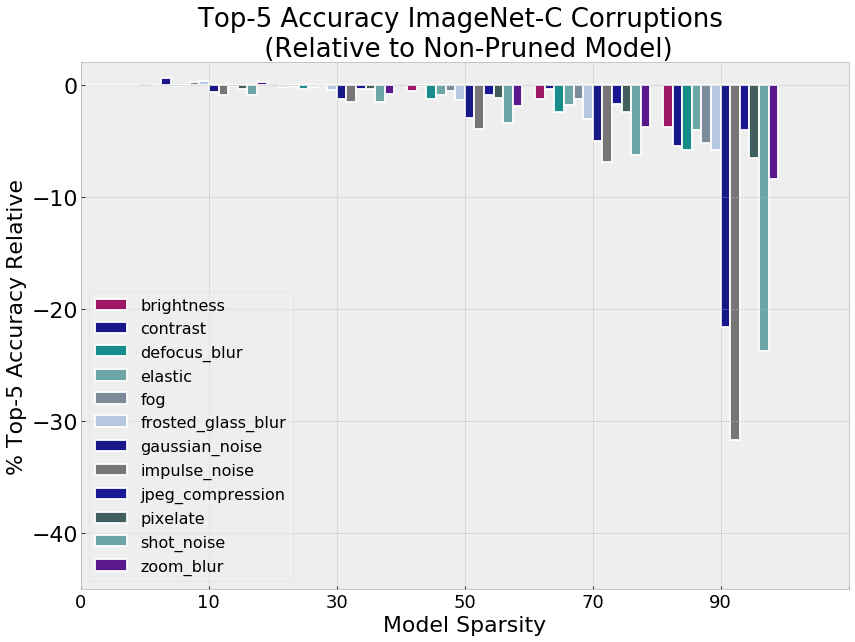}
  \caption{High levels of compression amplify sensitivity to distribution shift. \textbf{Left:} Change to top-1 normalized recall of a pruned model \textbf{relative} to a non-pruned model on ImageNet-C (all corruptions). \textbf{Right:} Change to top-5 normalized recall of a pruned model \textbf{relative} to a non-pruned model on ImageNet-C (all corruptions). We measure the top-1 test-set performance on a subset of ImageNet-C corruptions of a pruned model \textbf{relative} to the non-pruned model on the same corruption.}
\label{fig:performance_imagenet_c_imagenet_a_all} \end{center} 
\end{figure*}

\textbf{ImageNet-A Extended Results} ImageNet-A is a curated test set of $7,500$ natural adversarial images designed to produce drastically low test accuracy. We find that the sensitivity of pruned models to ImageNet-A mirrors the patterns of degradation to ImageNet-C and sets of PIEs. As pruning increases, top-1 and top-5 accuracy further erode, suggesting that pruned models are more brittle to adversarial examples. Table \ref{appendix:ImageNet_A_all_classes} includes relative and absolute sensitivity at all levels of compression considered.

\begin{table*}[ht!]
  \centering
\begin{tabular}{ccccc}
\toprule
 \multicolumn{5}{l}{\textbf{ImageNet Robustness to ImageNet-A Corruptions (By Level of Pruning)}} \\
 \midrule
\textbf{Pruning Fraction} & \textbf{Top-1 }  & \textbf{Top-5} & \textbf{Top-1 Norm} &  \textbf{Top-5 Norm} \\
\midrule
             0.0 &            0.89 &            7.56 &                 0.00 &                 0.00 \\
            10.0 &            0.85 &            7.53 &                -4.04 &                -0.39 \\
            30.0 &            0.76 &            7.21 &               -14.33 &                -4.62 \\
            50.0 &            0.62 &            6.53 &               -30.54 &               -13.65 \\
            70.0 &            0.51 &            5.83 &               -42.63 &               -22.96 \\
            90.0 &            0.36 &            4.47 &               -59.80 &               -40.96 \\
\bottomrule
\end{tabular}
 \caption{Pruned models are more sensitive to natural adversarial images. ImageNet-A is a curated test set of $7,500$ natural adversarial images designed to produce drastically low test accuracy. We compute the absolute performance of models pruned to different levels of sparsity on ImageNet-A (Top-1 and Top-5) as well as the normalized performance relative to a non-pruned model on ImageNet-A.}\label{appendix:ImageNet_A_all_classes} 
\end{table*}

For each robustness benchmark and level of pruning that we evaluate, we average model robustness over $5$ models independently trained from random initialization.

\textbf{ImageNet-C Extended Results} ImageNet-C \citep{hendrycks2018benchmarking} is an open source data set that consists of algorithmic generated corruptions (blur, noise) applied to the ImageNet test-set. We compare top-1 accuracy given inputs with corruptions of different severity. As described by the methodology of \citet{hendrycks2018benchmarking}, we compute the corruption error for each type of corruption by measuring model performance rate across five corruption severity levels (in our implementation, we normalize the per-corruption error by the performance of the non-compressed model on the same corruption). 

ImageNet-C corruption substantially degrades mean top-1 accuracy of pruned models relative to non-pruned. As seen in Fig.\ref{fig:performance_imagenet_c_imagenet_a_all}, this sensitivity is amplified at high levels of pruning, where there is a further steep decline in top-1 accuracy. Unlike the main body, in this figure we visualize all corruption types considered. Sensitivity to different corruptions is remarkably varied, with certain corruptions such as Gaussian, shot an impulse noise consistently causing more degradation. We include a visualization for a larger sample of corruptions considered in Table \ref{appendix:ImageNet_C_all_classes}.

\end{document}